\documentclass[preprint,12pt]{elsarticle}

\usepackage{amssymb}
\usepackage{microtype}
\usepackage{graphicx}
\usepackage[colorlinks=true,
            linkcolor=blue,
            citecolor=blue,
            urlcolor=blue]{hyperref}
\usepackage{algorithm}
\usepackage{amsmath}
\usepackage{multirow}
\usepackage{booktabs}
\usepackage[section]{placeins} 
\usepackage{algorithm}
\usepackage{algpseudocode}
\usepackage{float}

\newcommand{\etal}{\textit{et al.}}

\journal{Nuclear Physics B}

\begin{document}

\begin{frontmatter}



\title{Enlight: Fast Low-Light Image Enhancement via Multi-Objective Optimization and Shadow-Aware Refinement}





\author[a,b]{Nirjhor Datta}
\author[a]{M. Sohel Rahman}

\affiliation[a]{organization={Department of Computer Science and Engineering},
  addressline={Bangladesh University of Engineering and Technology},
  city={Dhaka},
  postcode={1205},
  country={Bangladesh}}

\affiliation[b]{organization={Department of Computer Science and Engineering},
  addressline={BRAC University},
  city={Dhaka},
  postcode={1205},
  country={Bangladesh}}

\begin{abstract}
We present ENLIGHT, a fast and training free framework for low-light image enhancement based on direct optimization of a perceptual objective. Unlike deep learning approaches that require large scale training data and supervision, ENLIGHT operates in a zero-shot manner by optimizing image quality at inference time.
The method employs a two stage global to local optimization strategy. In the first stage, ENLIGHT performs global illumination adjustment to improve visibility while maintaining structural consistency and avoiding excessive noise enhancement. In the second stage, a shadow aware refinement selectively improves low-intensity regions through masked local optimization, enhancing visibility without overexposure. To balance quality and efficiency, we introduce two modes: \emph{Fast}, which uses a multi-objective formulation combining entropy, gradient preservation, and noise regularization, and \emph{Ultrafast}, which reduces computational cost via a lightweight approximation of the same objective. The framework is optimizer agnostic and supports both evolutionary and lightweight local search methods. Experiments on BAID, Backlit300, LIME, MEF, NPE, and DICM demonstrate that ENLIGHT achieves competitive perceptual quality (MUSIQ, NIQE, BRISQUE) with significantly lower inference time. Qualitative results further show improved contrast, preserved structural details, and controlled noise amplification, making ENLIGHT a practical and interpretable alternative to learning based methods.
\end{abstract}




\end{frontmatter}

\section{Introduction}

Low-light image enhancement (LLIE) aims to restore visual fidelity and interpretability in images captured under poor illumination, where visibility degradation, noise, and color distortion adversely affect both human perception and downstream vision tasks~\cite{guo2016lime,guo2020zero,zhang2019kindling}. 
Classical approaches, including retinex-based illumination decomposition~\cite{land1971lightness,fu2016weighted,guo2016lime} and multi-exposure fusion~\cite{ma2015multi,li2018multi}, improve brightness and contrast but often struggle under spatially varying lightning conditions, leading to artifacts such as halo effects and color inconsistency.

Recent deep learning methods~\cite{wei2018deep,wu2022uretinex,jiang2021enlightengan} achieve strong performance by learning complex mappings from low-light to normal-light images. 
However, these approaches typically rely on large scale paired datasets and extensive training, limiting their adaptability to unseen conditions and reducing interpretability.

To address data dependence, zero-reference methods such as
Zero-DCE~\cite{guo2020zero}, Zero-DCE++~\cite{li2021learning},
RUAS~\cite{liu2021retinex}, and SCI~\cite{ma2022toward}
perform enhancement without paired supervision. However, these methods still rely on implicit neural optimization,
often exhibit instability under diverse lightning conditions,
and lack interpretability in relating enhancement decisions to visual information.

In parallel, no reference quality metrics such as BRISQUE~\cite{mittal2012no}, NIQE~\cite{mittal2013making}, and MUSIQ~\cite{ma2022toward} have emerged as important indicators of visual quality. 
However, existing LLIE approaches often incorporate these cues indirectly or inconsistently, leading to suboptimal trade-offs between contrast enhancement, structural preservation, and noise suppression. 
In particular, aggressive enhancement in dark regions frequently results in amplified noise and loss of local consistency.

In this work, we propose \textbf{ENLIGHT}, a training-free framework that formulates low-light enhancement as a direct optimization problem over a small set of interpretable global parameters. 
Instead of learning pixel-wise mappings, ENLIGHT optimizes brightness, contrast, and gamma using a perceptual objective that balances entropy maximization, structural consistency, and noise regularization.

A key aspect of the proposed method is a two-stage global-to-local optimization strategy. 
The first stage performs global enhancement to correct overall illumination, while the second stage introduces a \emph{shadow-aware refinement} that selectively improves low-intensity regions through masked local optimization. 
This design allows ENLIGHT to enhance dark regions without overexposing well-lit areas.

Furthermore, the framework is \emph{optimizer-agnostic} and supports both evolutionary and lightweight optimization strategies, enabling a flexible trade-off between accuracy and efficiency. 
We introduce two operating modes: \textit{Fast}, which performs full multi-objective optimization, and \textit{Ultrafast}, which reduces computational cost through a simplified objective and reduced search budget.

Extensive experiments on several well-known low-light image enhancement datasets, including BAID, Backlit300, LIME, MEF, NPE, and DICM, demonstrate that ENLIGHT achieves competitive perceptual quality compared to both classical and learning-based methods, while maintaining significantly lower inference time. 
In addition, qualitative analysis shows improved contrast, preserved edge structure, and controlled noise amplification, highlighting the interpretability and robustness of the proposed approach.

Our contributions are summarized as follows:
\begin{enumerate}
    \item We propose \textbf{ENLIGHT}, a training-free framework for low-light image enhancement based on direct optimization of interpretable global parameters.
    \item We design a multi-objective formulation that jointly balances entropy maximization, structural preservation, and noise suppression.
    \item We introduce a two-stage global-to-local optimization strategy with shadow-aware refinement for improved enhancement of low-intensity regions.
    \item We develop an optimizer-agnostic framework supporting evolutionary and lightweight optimization methods, with \textit{Fast} and \textit{Ultrafast} modes for efficient inference.
    \item We demonstrate through extensive experiments that ENLIGHT achieves strong perceptual quality and generalization while operating at low computational cost.
\end{enumerate}

\section{Related Work}
\label{sec:related}

Low-light image enhancement (LLIE) has been extensively explored through a variety of paradigms spanning physics-based decomposition, deep learning, and perceptual optimization. 
This section reviews representative developments across these domains and outlines the distinctive position of our proposed \textbf{Enlight} framework.

\subsection{Retinex and Fusion-Based Enhancement Methods}
Traditional LLIE approaches are primarily grounded in the Retinex theory ~\cite{land1971lightness}, which decomposes an image into illumination and reflectance components.
  Supraja \textit{et al.}~\cite{guo2016lime} introduced the Low-Light Image Enhancement via Illumination Map Estimation (LIME) framework, which focuses on estimating a pixel-wise illumination map to perform non-uniform brightness correction and detail restoration. Their method first computes an initial illumination estimate by taking the maximum value across RGB channels for each pixel and then refines it to avoid over-saturation. To further enhance visibility and suppress noise, Block Matching 3D (BM3D) filtering and Non-Local Means (NLM) denoising are applied, demonstrating that illumination estimation coupled with spatial filtering yields superior perceptual quality compared with traditional global approaches. 

Earlier enhancement methods such as histogram equalization and contextual variational contrast enhancement improved global contrast but often over-amplified noise or distorted color balance, particularly under heterogeneous lighting conditions. In contrast, LIME achieves adaptive enhancement by handling non-uniform illumination and preserving natural image structures. The study concludes that accurate illumination estimation can substantially improve downstream computer-vision tasks such as edge detection, feature matching, and object recognition by providing high-visibility inputs.

Another work  ~\cite{fu2016weighted} proposed the Simultaneous Reflectance and Illumination Estimation (SRIE)  model, a weighted variational Retinex framework that jointly estimates both reflectance and illumination from a single image. Unlike conventional Retinex-based methods that rely on logarithmic transformations and often lead to over-smoothed reflectance, SRIE revisits the variational formulation and introduces weighted regularization terms to better preserve structural details. By analyzing the side effects of the logarithmic transform, the authors demonstrate that using it directly in the regularization term suppresses gradient magnitudes in bright regions, causing loss of fine textures. To address this, SRIE weights the total variation priors with the exponential of the estimated components, enabling simultaneous estimation of illumination and reflectance through an alternating minimization scheme solved via the ADMM algorithm. Experimental evaluations show that SRIE achieves superior enhancement quality, better texture preservation, and lower noise amplification compared with classical Retinex models such as those by Kimmel~\cite{kimmel2003variational} and Ng~\cite{ng2011total}. The method demonstrates high naturalness and robustness across various illumination conditions, making it a foundational contribution to modern low-light image enhancement.

Although these models are interpretable, they struggle to generalize under varying illumination and often introduce color distortion or halo artifacts. 
Fusion-based techniques, such as MFIF~\cite{li2018multi} and MEF~\cite{ma2015multi}, attempt to combine multiple exposure-corrected versions of the same image to restore balanced luminance, yet they are computationally expensive and require precise tone mapping heuristics.In contrast, \textsc{Enlight} does not rely on reflectance decomposition or fusion heuristics. Instead, it directly optimizes a perceptual objective in the image domain, combining entropy, gradient consistency, and noise regularization to guide illumination adjustment. 


\subsection{Learning-Based Low-Light Enhancement}
The emergence of deep neural networks has dramatically advanced LLIE performance, with supervised architectures like RetinexNet~\cite{wei2018deep}, URetinex-Net~\cite{wu2022uretinex}, KinD~\cite{zhang2019kindling}, and EnlightenGAN~\cite{jiang2021enlightengan} setting strong benchmarks through data-driven learning.

The progression of deep learning based low-light image enhancement (LLIE) frameworks has transitioned from physically interpretable Retinex formulations to unpaired and self-regularized adversarial learning. Wei \textit{et al.}~\cite{wei2018deep} pioneered this line with \textbf{Retinex-Net}, an end-to-end trainable model that decomposes an image into illumination and reflectance through a Decom-Net and subsequently adjusts illumination via an Enhance-Net with multi-scale concatenation. Trained on the LOL dataset, Retinex-Net achieved illumination-aware enhancement while preserving structural details and reducing noise. Building upon this foundation, Zhang \textit{et al.}~\cite{zhang2019kindling} proposed \textbf{KinD (Kindling the Darkness)}, a practical Retinex-inspired system integrating three cooperative sub-networks for decomposition, reflectance restoration, and illumination adjustment. KinD enables user-controllable exposure scaling and jointly suppresses noise and color distortion, producing natural brightness restoration under real-scene conditions. Later, Wu \textit{et al.}~\cite{wu2022uretinex} introduced \textbf{URetinex-Net}, a deep unfolding Retinex network that bridges optimization-based interpretability and deep feature learning. By unrolling the Retinex decomposition into iterative trainable stages with adaptive priors, URetinex-Net achieves superior illumination consistency, generalization, and structural fidelity. 

In contrast to these supervised frameworks, Jiang \textit{et al.}~\cite{jiang2021enlightengan} presented \textbf{EnlightenGAN}, an unpaired GAN-based model that dispenses with paired supervision through global-local discriminators and a self-regularized perceptual constraint. This design preserves content semantics while achieving adaptive enhancement of dark regions. Collectively, these models mark the evolution from paired, physically grounded Retinex learning toward unpaired, generative, and perceptually guided enhancement, balancing interpretability, flexibility, and visual realism across diverse lighting conditions


While prior deep learning models often rely on illumination decomposition or adversarial mappings, they may not explicitly align optimization with perceptual quality measures. The proposed \textbf{Enlight} framework addresses this by introducing a two-stage optimization pipeline that combines global entropy-driven enhancement with shadow-aware local refinement. 

Unlike data-driven approaches, \textsc{Enlight} estimates enhancement parameters directly at test time through a perceptual objective, enabling adaptive and interpretable adjustment without requiring training data. This design provides a practical alternative that balances detail preservation, exposure correction, and noise suppression across diverse lighting conditions.

\subsection{Perceptual and Optimization-Driven Methods}

Recent research has increasingly focused on perceptually grounded low-light image enhancement (LLIE), emphasizing human-aligned quality measures such as BRISQUE~\cite{mittal2012no}, NIQE~\cite{mittal2013making}, and MUSIQ~\cite{ma2022toward}. These metrics have motivated a new generation of optimization-driven frameworks that seek to maximize perceptual fidelity without explicit supervision. Classical approaches, including Retinex-inspired entropy maximization~\cite{zhu2020eemefn}, unsupervised gradient-based illumination tuning~\cite{ying2017new}, and probabilistic variational enhancement~\cite{fu2015probabilistic}, demonstrate the potential of physics-constrained optimization to jointly improve visibility and preserve structural integrity. However, their reliance on manually designed objective functions and globally uniform enhancement parameters often limits adaptability across spatially heterogeneous illumination conditions. 

To address these challenges, \textsc{Enlight} formulates low-light enhancement as an optimization problem guided by a perceptual objective. Specifically, entropy, gradient consistency, and Laplacian-based noise estimation jointly regulate illumination adjustment. By explicitly balancing exposure enhancement with noise suppression, the proposed framework provides a simple and interpretable formulation that generalizes across diverse imaging conditions.

\paragraph{\textbf{Summary.} }
In summary, while Retinex-based methods rely on illumination decomposition and deep learning approaches depend on data-driven mappings, \textbf{Enlight} formulates low-light enhancement as a training-free optimization problem. By directly optimizing a perceptual objective, it combines the interpretability of model-based methods with the adaptability of data-driven approaches.

The proposed framework achieves competitive visual quality while maintaining near real-time performance. Its formulation provides explicit control over brightness, contrast, and noise, enabling stable and consistent enhancement across diverse imaging conditions. Overall, \textsc{Enlight} offers a practical balance between efficiency, interpretability, and perceptual quality.

\begin{figure*}[t]
    \centering
    \includegraphics[width=0.85\textwidth]{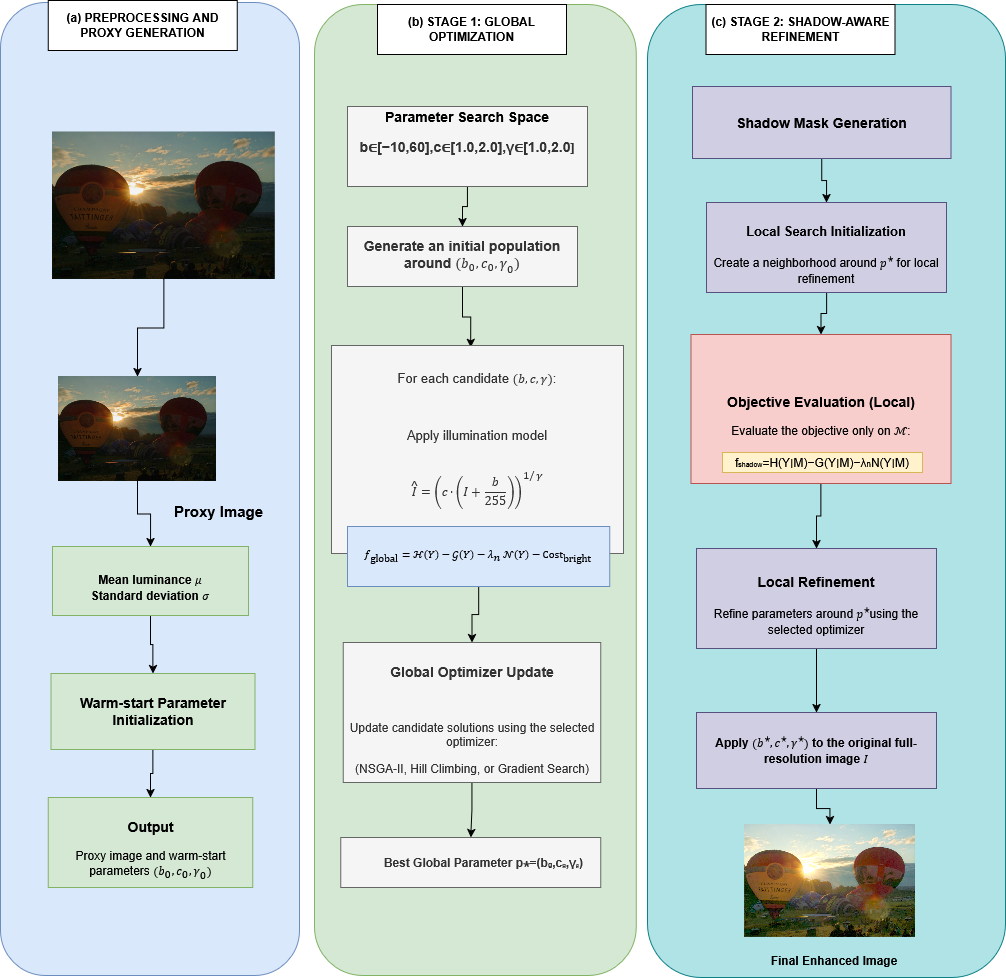}
    \caption{Top level overview of the framework}
    \label{fig:overview}
\end{figure*}

\section{Proposed Method}

\subsection{Framework Overview}
The proposed ENLIGHT framework formulates low-light image enhancement as a training-free optimization problem over a compact set of interpretable parameters. Instead of learning pixel-wise mappings, ENLIGHT directly searches for optimal brightness, contrast, and gamma parameters using a perceptually guided objective. The framework follows a global-to-local optimization strategy, where global illumination is first corrected and subsequently refined through a shadow-aware local stage. To ensure efficiency, optimization is performed on a downscaled proxy image with a warm-start initialization, and supports multiple optimizers under \emph{Fast} and \emph{Ultrafast} configurations. This design enables ENLIGHT to achieve a balance between perceptual quality, interpretability, and computational efficiency. The overall pipeline is summarized in Fig.~\ref{fig:overview} and Algorithm~\ref{alg:enlight}.

\subsection{Parametric Illumination Transform}
Given an input image $I \in [0,1]^{H \times W \times 3}$, Enlight employs a three-parameter illumination transform defined as
\begin{equation}
    \hat{I} = \left(c \cdot (I + b/255)\right)^{1/\gamma},
\end{equation}
where $b$ denotes a brightness offset, $c$ a contrast scaling factor, and $\gamma$ a nonlinear gamma exponent. To ensure stability and physical interpretability, the parameters are constrained to 
\begin{equation}
    b \in [-10,\,60],\quad c \in [1.0,\,2.0],\quad \gamma \in [1.0,\,2.0].
\end{equation}

The parameter ranges were selected to cover typical brightness, contrast, and gamma adjustments encountered in low-light enhancement while avoiding unrealistic transformations. In particular, the brightness range allows sufficient illumination correction without excessive saturation, whereas the contrast and gamma bounds restrict the search to monotonic tone-mapping functions that preserve relative intensity ordering and reduce the risk of color distortion or over-enhancement. These bounds also limit the optimization space, improving convergence efficiency during evolutionary search.

\subsection{Perceptual Objective}
\label{sec:global}
To guide optimization without supervision, Enlight evaluates each candidate solution by computing a proxy measure designed to correlate with perceptual visibility. Let $Y$ denote the grayscale luminance of the enhanced image. Three complementary criteria are employed:
\begin{itemize}
    \item \textbf{Entropy} $\mathcal{H}(Y)$ encourages distributional spread and mitigates global underexposure.
    \item \textbf{Gradient consistency} $\mathcal{G}(Y)$ penalizes deviations from the input image gradients (Laplacian and Sobel), encouraging structural consistency between the input and enhanced images.
    \item \textbf{Laplacian variance} $\mathcal{N}(Y)$ acts as a noise proxy to suppress over-enhancement.
\end{itemize}

The global proxy objective is defined as
\begin{equation}
f_{\text{global}} = \mathcal{H}(Y) - \mathcal{G}(Y) - \lambda_{n}\mathcal{N}(Y) - \text{Cost}_{\text{bright}},
\end{equation}
 where $\text{Cost}_{\text{bright}}$ penalizes deviations from a desired luminance range and $\lambda_{n}$ controlling the noise penalty..

To prevent unrealistic exposure, the mean luminance $\mu(Y)$ is constrained via:
\begin{equation}
\label{eq:brightnessrange}
\text{Cost}_{\text{bright}} =
\begin{cases}
 \alpha(L - \mu(Y))^{+}, & \mu(Y) < L,\\
 \alpha(\mu(Y) - U)^{+}, & \mu(Y) > U,
\end{cases}
\end{equation}
where $L{=}0.35$, $U{=}0.70$ define a conservative target range for global brightness, $(x)^{+}=\max(x,0)$, and $\alpha$ controls the penalty strength. This range discourages both under-enhancement and overexposure while allowing sufficient flexibility for different low-light scenes.

\subsection{Shadow-Aware Local Refinement}
\label{sec:shadow}

To improve under-exposed regions, ENLIGHT introduces a shadow-aware local refinement stage. Given the grayscale luminance map $Y$, a shadow mask is defined as
\begin{equation}
    \mathcal{M} = \{x : Y(x) < \tau\}, \quad \tau = 0.28.
\end{equation}
The threshold $\tau$ is selected empirically as a conservative boundary for identifying strongly under-exposed pixels. Preliminary sensitivity checks showed that nearby values produce similar perceptual scores, suggesting that the refinement stage is not highly sensitive to the exact threshold.

The objective is then evaluated only over $\mathcal{M}$, enabling masked local optimization focused on shadow regions:
\begin{equation}
    f_{\text{shadow}} =
    \mathcal{H}(Y|\mathcal{M}) -
    \mathcal{G}(Y|\mathcal{M}) -
    \lambda_n \mathcal{N}(Y|\mathcal{M}).
\end{equation}
This masked formulation guides the optimizer toward better recovery of low-intensity regions while reducing the risk of over-enhancing already well-exposed areas. The final enhancement is still applied globally using the optimized parameters; the mask is used only to guide the objective during optimization.

\subsection{Warm-Start Parameter Initialization}

To reduce search overhead, ENLIGHT employs a luminance-driven warm-start heuristic based on simple image statistics. Let $\mu$ and $\sigma$ denote the mean and standard deviation of the input grayscale image. The initial parameters are defined as:
\begin{align}
    b_0 &= 0.85(0.5 - \mu)\cdot 255, \\
    c_0 &= 
    \begin{cases}
        1.25 + 0.5(0.12 - \sigma), & \sigma < 0.12, \\
        1.05 + 0.3(0.2 - \sigma),  & \text{otherwise},
    \end{cases} \\
    \gamma_0 &= 
    \begin{cases}
        1.35, & \mu < 0.40, \\
        1.15, & \mu < 0.55, \\
        1.05, & \text{otherwise}.
    \end{cases}
\end{align}

These coefficients are derived from empirical observations of low-light image statistics and are designed to provide a stable and meaningful initialization. In particular, the brightness offset $b_0$ shifts the mean luminance toward a mid-range target ($\mu \approx 0.5$), while the contrast parameter $c_0$ increases for low-variance inputs to enhance visibility in flat regions. The gamma parameter $\gamma_0$ is adapted based on overall brightness, enabling stronger nonlinear correction for darker images.

Importantly, the warm-start does not determine the final solution but serves as an initialization that reduces the effective search space and accelerates convergence. The subsequent optimization stage refines these parameters based on the perceptual objective. In practice, this heuristic consistently guides the optimizer toward high-quality solutions across datasets without requiring dataset-specific tuning, improving efficiency in both \emph{Fast} and \emph{Ultrafast} modes.
\subsection{Optimization Procedure}

The parameter triplet $(b,c,\gamma)$ is optimized using a two-stage, proxy-guided search strategy that is independent of any specific optimizer. The first stage performs global optimization using the objective $f_{\text{global}}$ (Sec.\ref{sec:global}), while the second stage refines the solution using a shadow-aware objective $f_{\text{shadow}}$ (Sec.\ref{sec:shadow}).

\textbf{Algorithm~\ref{alg:enlight}} summarizes the complete optimization pipeline used in \textsc{Enlight}. The process begins with a warm-start estimate $(b_0,c_0,\gamma_0)$ derived from global brightness and contrast statistics of the input image. This initialization provides a stable anchor that prevents convergence to degenerate solutions and accelerates optimization.

An initial population (or candidate set) is constructed by perturbing the warm-start parameters, along with a small number of boundary samples to ensure sufficient exploration. The optimization then proceeds in two stages.

In the first stage, global optimization is performed for $T_1$ iterations. For each candidate parameter set $p$, the objective $f_{\text{global}}$ evaluates entropy maximization, gradient consistency, brightness constraints, and optional noise regularization. The candidate set is iteratively updated using an optimization operator (e.g., NSGA-II, hill climbing, or gradient-based updates), enabling flexible trade-offs between exploration and efficiency. After $T_1$ iterations, the best solution $p^\star$ is selected.

In the second stage, a shadow-aware refinement is applied. A binary mask $\mathcal{S}$ is constructed by thresholding the grayscale intensity of the enhanced image to identify low-intensity (shadow) regions. A local neighborhood is then generated around $p^\star$ via small perturbations. Each candidate is evaluated using $f_{\text{shadow}}$, which focuses exclusively on the masked region $\mathcal{S}$ to improve visibility in dark areas while controlling noise amplification. The candidate set is further refined for $T_2$ iterations using the same optimizer.

Finally, the optimal parameter set $(b,c,\gamma)$ is applied to the full-resolution image to produce the enhanced output. This global-to-local design enables \textsc{Enlight} to correct overall illumination while selectively enhancing shadow regions, resulting in improved perceptual quality with low computational overhead.
\begin{algorithm}[t]
\caption{ENLIGHT Optimization Pipeline}
\label{alg:enlight}
\begin{algorithmic}[1]
\Require Input image $I$, population size $N$, iterations $(T_1,T_2)$, mode $\in \{\text{Fast}, \text{Ultrafast}\}$
\State Adjust $(N, T_1, T_2)$ based on mode
\State Initialize $(b_0,c_0,\gamma_0)$ using warm-start rules
\State Construct initial candidate set $\mathcal{P}_0$
\For{$t \gets 1$ to $T_1$} \Comment{Global optimization}
    \ForAll{$p \in \mathcal{P}_t$}
        \State Evaluate $f_{\text{global}}(p)$
    \EndFor
    \State Update $\mathcal{P}_t$ using optimizer (e.g., NSGA-II / Hill / Gradient)
\EndFor
\State $p^\star \gets \arg\max f_{\text{global}}$
\State Generate shadow mask $\mathcal{S}$ using intensity threshold $\tau$
\State Construct local neighborhood $\mathcal{N}(p^\star)$
\For{$t \gets 1$ to $T_2$} \Comment{Shadow refinement}
    \ForAll{$q \in \mathcal{N}(p^\star)$}
        \State Evaluate $f_{\text{shadow}}(q)$ on $\mathcal{S}$
    \EndFor
    \State Update neighborhood using optimizer
\EndFor
\State Select optimal $(b,c,\gamma)$
\State Render enhanced image $\hat{I}$ using $(b,c,\gamma)$
\Return $\hat{I}$

\end{algorithmic}
\end{algorithm}
\subsection{Computational Complexity}

Let $N$ denote the number of candidate solutions, $T_1$ and $T_2$ the number of iterations in the global and local stages, respectively, and $H_s \times W_s$ the resolution of the proxy image. Each evaluation involves convolution, histogram estimation, and point-wise operations, resulting in a per-candidate cost of
\begin{equation}
    \mathcal{O}(H_s W_s).
\end{equation}
The total complexity is therefore
\begin{equation}
    \mathcal{O}\big((T_1 + T_2)\, N\, H_s\, W_s\big).
\end{equation}

In practice, this cost is significantly reduced due to three factors. First, optimization is performed on a downscaled proxy image ($H_s, W_s \ll H, W$), which preserves global statistics while reducing computation. Second, GPU-native tensor operations enable efficient parallel evaluation of candidates. Third, the warm-start initialization reduces the effective search space and accelerates convergence.

Furthermore, the proposed \emph{Fast} and \emph{Ultrafast} modes adjust $(N, T_1, T_2)$ to trade off between quality and runtime. In particular, the \emph{Ultrafast} configuration uses a reduced search budget, enabling near real-time performance while maintaining comparable perceptual quality.
\subsection{Implementation Details}
{\sloppy
All experiments were performed on a high-performance workstation equipped with
an Intel Core i9-14900K CPU (24 cores, 32 threads) and an NVIDIA RTX A6000 GPU
with 48~GB of VRAM and 64~GB of DDR5 system memory. Storage consisted of
a 1~TB NVMe SSD. A 1000~W power supply ensured stable operation during
extended experimental runs.
\par}

\subsection{FAST and ULTRAFAST Operating Modes}
To accommodate different runtime constraints while maintaining perceptual fidelity, Enlight provides two computational modes: \emph{FAST} and \emph{ULTRAFAST}. Both modes share the same two-stage evolutionary pipeline, proxy objective, and full-resolution rendering. The distinction lies solely in the computational budget allocated to the proxy-level optimization.

\paragraph{FAST Mode}
FAST mode employs the full optimization schedule, consisting of $T_{1}=16$ global generations and $T_{2}=4$ shadow-refinement generations, with a population size of $N=40$. All individuals are evaluated at a proxy resolution of 256 pixels on the image’s longer side. This configuration balances perceptual accuracy and computational cost, and serves as the default operating point for most applications.

\paragraph{ULTRAFAST Mode}
ULTRAFAST mode automatically reduces the evolutionary budget by scaling the parameters from FAST mode according to
\begin{equation}
    T_{1}^{\mathrm{UF}} = \max(4,\; T_{1}/4), \quad
    T_{2}^{\mathrm{UF}} = \max(2,\; T_{2}/2),
\end{equation}
\begin{equation}
    N^{\mathrm{UF}} = \max(16,\; N/2),
\end{equation}
while retaining the same proxy resolution. This results in approximately a $4\times$ reduction in optimization iterations and a $2\times$ reduction in population size. Despite the smaller search budget, warm-start initialization (Sec.~III-E) ensures stable convergence, enabling high-quality enhancement at significantly lower runtime.

Both modes yield identical illumination transforms and use the same fitness formulation; the difference lies only in the depth of the evolutionary search. FAST is suitable for applications requiring maximal perceptual fidelity, whereas ULTRAFAST provides near real-time performance with minimal degradation.

\section{Experiments}
We evaluate ENLIGHT on multiple benchmark datasets to assess perceptual quality, efficiency, and robustness across different illumination conditions.

\subsection{Datasets}
\noindent\textbf{Unsupervised Datasets}
We assess the performance of our method on \emph{unpaired} low-light images. 
Because our approach does not rely on paired ground-truth images, it is naturally compatible with real-world low-light enhancement scenarios. 
To validate this capability, we evaluate our method on five widely used unpaired LLIE datasets: 
DICM~\cite{lee2013contrast}, LIME~\cite{guo2016lime}, MEF~\cite{ma2015perceptual}, NPE~\cite{wang2013naturalness}, and VV~\cite{vonikakis2018evaluation}. 
A concise overview of these datasets is provided in Table~\ref{tab:datasets}.

\begin{table}[H]
\centering
\small
\caption{Summary of Unpaired Low-Light Image Datasets Used in Our Experiments}
\label{tab:datasets}
\begin{tabular}{|l|c|p{3.8cm}|c|}
\hline
\textbf{Dataset} & \textbf{Images} & \textbf{Description} & \textbf{Format} \\
\hline
\textbf{DICM} & 69 & Real-world low-light scenes including night streets, vehicle lighting, underwater flowers, dark indoor rooms, and shadow-heavy environments. & JPG \\
\hline
\textbf{LIME} & 10 & Natural scenes captured under insufficient illumination, often with strong global underexposure. & BMP \\
\hline
\textbf{MEF} & 17 & A mixture of indoor and outdoor natural scenes, typically high-resolution, captured in low-light or uneven illumination. & PNG \\
\hline
\textbf{NPE} & 8 & Outdoor images with non-uniform illumination and severe visibility degradation. & JPG \\
\hline
\textbf{VV} & 24 & Daytime outdoor environments that appear dim due to poor exposure (urban streets, vehicles, buildings). & JPG \\
\hline
\end{tabular}
\end{table}
\noindent\textbf{CLIP-LIT\cite{liang2023iterative} Benchmark}
Following the evaluation protocol of CLIP-LIT\cite{liang2023iterative}, we adopt the BAID
and Backlit300 datasets to assess the performance of our method under
challenging backlit conditions. The BAID dataset contains 748 real backlit
images, divided into 380 training images and 368 testing images. Since our
approach is fully training-free, we use only the 368 backlit test images for
quantitative and qualitative evaluation. To further examine generalization, we
also evaluate on the Backlit300 collection introduced in CLIP-LIT\cite{liang2023iterative}, which
consists of 305 diverse backlit photographs gathered from the Internet, Pexels,
and Flickr. These datasets include a wide range of lighting conditions,
illumination directions, and scene types, making them suitable for stress-testing
the robustness of enhancement models. All images from BAID and Backlit300 used
in our experiments are publicly available.

\textbf{MIT-5K Dataset}To further assess the generalization ability of our method on professional
photography, we evaluate \textsc{Enlight} on the MIT Adobe FiveK dataset \cite{bychkovsky2011learning},
which contains 5,000 high-resolution RAW photographs retouched by five expert
photographers to produce studio-quality target images. Although the dataset is
typically used to train supervised enhancement models, our approach does not
utilize any of the provided ground-truth retouched images during optimization.
Instead, we process the MIT5K photos directly and compare our outputs against
the expert-adjusted references for quantitative analysis. 
\subsubsection{Evaluation Metrics}

We evaluate enhancement quality using both no-reference and full-reference image quality assessment metrics. Specifically, we employ BRISQUE~\cite{mittal2012no}, NIQE~\cite{mittal2013making}, MUSIQ~\cite{ma2022toward}, Peak Signal-to-Noise Ratio (PSNR), and Structural Similarity Index Measure (SSIM). These metrics collectively assess perceptual quality, structural fidelity, statistical naturalness, and reconstruction consistency.

BRISQUE and NIQE are no-reference metrics based on natural scene statistics, where lower values indicate better perceptual quality. BRISQUE evaluates locally normalized luminance statistics associated with visual distortions, while NIQE measures deviations from statistical regularities observed in natural images without requiring supervised training. MUSIQ is a transformer-based perceptual quality metric designed to capture high-level visual fidelity and human perceptual preference, where higher values correspond to better image quality.

For reference-based evaluation, we additionally compute PSNR and SSIM. PSNR measures reconstruction fidelity between the enhanced image and the reference image, and is defined as
\begin{equation}
\text{PSNR} = 10 \log_{10}\left(\frac{\text{MAX}^2}{\text{MSE}}\right),
\end{equation}
where MAX denotes the maximum possible pixel value and MSE represents the Mean Squared Error between the enhanced and reference images. Higher PSNR values indicate better reconstruction quality.

SSIM evaluates structural similarity by jointly comparing luminance, contrast, and structural information:
\begin{equation}
\text{SSIM}(x,y)=
\frac{(2\mu_x\mu_y + C_1)(2\sigma_{xy}+C_2)}
{(\mu_x^2+\mu_y^2+C_1)(\sigma_x^2+\sigma_y^2+C_2)},
\end{equation}
where $\mu_x$ and $\mu_y$ denote the mean intensities of images $x$ and $y$, $\sigma_x^2$ and $\sigma_y^2$ represent variances, $\sigma_{xy}$ is the covariance, and $C_1,C_2$ are small constants for numerical stability. Higher SSIM values indicate better structural preservation.

Together, these metrics provide complementary evaluation of contrast enhancement, structural preservation, perceptual realism, naturalness, and noise suppression, enabling comprehensive assessment of low-light enhancement performance.
\subsection{Optimizer Comparison and Design Justification}

We first evaluate the impact of different optimization strategies on both perceptual quality and computational efficiency. Specifically, we compare NSGA-II, hill climbing, and gradient-based local search under both \emph{Fast} and \emph{Ultrafast} configurations.

\begin{table}[H]
\centering
\small
\setlength{\tabcolsep}{5pt}
\renewcommand{\arraystretch}{1.2}
\begin{tabular}{|l|l|l|c|c|c|}
\hline
\textbf{Dataset} & \textbf{Optimizer} & \textbf{Mode} & \textbf{Time (s)} & \textbf{BRISQUE $\downarrow$} & \textbf{NIQE $\downarrow$} \\
\hline
\multirow{6}{*}{DICM}
& Hill  & Ultrafast & 0.14  & 18.69 & 3.63 \\
& NSGA-II & Ultrafast & 0.13  & 20.03 & 3.70 \\
& Local & Ultrafast & \textbf{0.118} & \textbf{17.07} & 3.66 \\
& Hill  & Fast      & 0.168 & 19.34 & 3.68 \\
& NSGA-II & Fast      & 0.700 & 21.77 & 3.73 \\
& Local & Fast      & 0.215 & 17.14 & \textbf{3.65} \\
\hline
\multirow{6}{*}{LIME}
& Local & Fast      & 0.23  & 22.78 & 4.32 \\
& Hill  & Fast      & 0.207 & 19.28 & 4.20 \\
& NSGA-II & Fast      & 0.613 & \textbf{18.60} & \textbf{4.18} \\
& NSGA-II & Ultrafast & 0.137 & 19.23 & 4.20 \\
& Hill  & Ultrafast & 0.165 & 19.44 & 4.20 \\
& Local & Ultrafast & \textbf{0.134} & 22.74 & 4.31 \\
\hline
\multirow{6}{*}{MEF}
& Local & Ultrafast & \textbf{0.119} & 13.90 & 3.61 \\
& Hill  & Ultrafast & 0.153 & 13.29 & 3.45 \\
& NSGA-II & Ultrafast & 0.124 & 13.57 & 3.44 \\
& Local & Fast      & 0.217 & 13.75 & 3.61 \\
& Hill  & Fast      & 0.219 & 13.62 & \textbf{3.41} \\
& NSGA-II & Fast      & 0.719 & \textbf{13.64} & 3.43 \\
\hline
\multirow{6}{*}{NPE}
& Local & Ultrafast & \textbf{0.120} & \textbf{17.72} & \textbf{3.66} \\
& Hill  & Ultrafast & 0.139 & 17.98 & 3.77 \\
& NSGA-II & Ultrafast & 0.137 & 18.23 & 3.77 \\
& Local & Fast      & 0.207 & \textbf{17.78} & 3.64 \\
& Hill  & Fast      & 0.167 & 18.03 & 3.76 \\
& NSGA-II & Fast      & 0.648 & 18.31 & 3.74 \\
\hline
\multirow{6}{*}{VV}
& Local & Ultrafast & \textbf{0.187} & 24.40 & 3.01 \\
& Hill  & Ultrafast & 0.239 & 22.19 & \textbf{2.72} \\
& NSGA-II & Ultrafast & 0.189 & \textbf{21.74} & \textbf{2.72} \\
& Local & Fast      & 0.286 & 24.41 & 3.02 \\
& Hill  & Fast      & 0.299 & 22.14 & \textbf{2.72} \\
& NSGA-II & Fast      & 0.715 & 22.32 & 2.73 \\
\hline
\end{tabular}
\caption{Comparison of optimization strategies under \emph{Fast} and \emph{Ultrafast} configurations. Lower BRISQUE and NIQE indicate better perceptual quality.}
\label{tab:optimizer_comparison}
\end{table}

As shown in Table~\ref{tab:optimizer_comparison}, lightweight optimizers consistently achieve significantly lower inference time. In particular, the \emph{Ultrafast} mode enables near real-time performance, with local search achieving the fastest runtime across datasets. In contrast, NSGA-II incurs higher computational cost due to population-based exploration.

The comparison reveals that no single optimizer dominates across all datasets. Local search achieves the best results on DICM and NPE, suggesting that the warm-start initialization often places the solution near a favorable region of the search space. In contrast, NSGA-II obtains the strongest perceptual quality on LIME and VV, where the optimization landscape may be more challenging due to greater variation in scene brightness and shadow structure. These results indicate that population-based search can provide additional robustness in difficult cases, while lightweight local methods remain attractive when computational efficiency is the primary objective.

Hill climbing and local search primarily exploit the search space, enabling fast convergence but limited exploration. In contrast, NSGA-II maintains population diversity and performs global exploration, allowing it to better navigate the objective landscape.

We observe that NSGA-II performs better on certain datasets (e.g., LIME, VV), while lightweight optimizers (hill/local) are competitive or superior on others (e.g., DICM, NPE). This behavior can be explained by differences in the underlying objective landscape.

Datasets with complex illumination patterns and stronger noise characteristics typically induce a more non-convex optimization landscape. In such cases, NSGA-II benefits from population-based exploration, enabling it to escape poor local optima and discover more balanced entropy–cost trade-offs.

In contrast, datasets with relatively smoother illumination distributions or simpler structures exhibit a more well-behaved objective surface. Under these conditions, local or hill-based methods can efficiently converge to high-quality solutions, often with significantly lower computational cost.

This highlights a fundamental trade-off between exploration and exploitation: NSGA-II provides robustness in complex scenarios, while lightweight optimizers offer efficiency when the optimization landscape is simpler.

Although lightweight local search methods occasionally achieve lower runtime or stronger performance on individual metrics, their behavior is less consistent across datasets and illumination conditions. In contrast, NSGA-II provides a more stable multi-objective optimization process by jointly balancing entropy enhancement, structural preservation, and noise suppression through population-based search. This results in more consistent perceptual quality across datasets, particularly under challenging illumination variations. Therefore, NSGA-II is adopted in the subsequent experiments as the primary optimizer for detailed evaluation and ablation analysis, while hill climbing and local search are retained as lightweight alternatives for efficiency-oriented settings.

The optimizer comparison is conducted under a unified and slightly simplified pipeline to ensure fair comparison across different solvers. In particular, the shadow-refinement stage uses a simplified masked objective. Subsequent experiments use the full NSGA-II configuration with the complete perceptual proxy, which may lead to slight differences in reported values.
\subsection{Quantitative Results on Unpaired Datasets}Table~\ref{tab:musiq_comparison} presents a detailed comparison of BRISQUE and NIQE 
scores across the five standard unpaired LLIE datasets used in our evaluation: 
DICM~\cite{lee2013contrast}, LIME~\cite{guo2016lime},MEF~\cite{ma2015perceptual}, NPE~\cite{wang2013naturalness}, and VV~\cite{vonikakis2018evaluation}. Our \textsc{Enlight} optimizer consistently 
produces competitive perceptual quality, with the FAST configuration achieving 
the strongest overall performance across multiple datasets. The ablation variants 
demonstrate that each component gradient consistency term, noise penalty, and 
shadow-refinement stage contributes meaningfully to output quality. Notably, the 
ULTRAFAST (no-noise) variant obtains the lowest or near-lowest distortion metrics on 
DICM whereas the ULTRAFAST (no-shadow) variant achieved the best result on MEF, NPE and VV, validating the effectiveness of our multi-objective proxy and 
the stability of the evolutionary search across diverse illumination conditions. Overall, our method 
achieved state of the art results on 4 out of 5 datasets.
\begin{table}[H]
\centering
\small
\caption{Comprehensive comparison of BRISQUE (B) and NIQE (N) scores ($\downarrow$ lower is better) across five unpaired LLIE datasets. Best results per column are in \textbf{bold}.}
\label{tab:big_llie_results}
\resizebox{\linewidth}{!}{%
\begin{tabular}{lcccccccccc}
\toprule
\multirow{2}{*}{\textbf{Method}} 
  & \multicolumn{2}{c}{\textbf{DICM}} 
  & \multicolumn{2}{c}{\textbf{LIME}} 
  & \multicolumn{2}{c}{\textbf{MEF}} 
  & \multicolumn{2}{c}{\textbf{NPE}} 
  & \multicolumn{2}{c}{\textbf{VV}} \\
\cmidrule(lr){2-3}\cmidrule(lr){4-5}\cmidrule(lr){6-7}\cmidrule(lr){8-9}\cmidrule(lr){10-11}
  & B & N  & B & N  & B & N  & B & N  & B & N \\
\midrule
KinD~\cite{zhang2019kindling}        & 48.72 & 5.15 & 39.91 & 5.03 & 49.94 & 5.47 & 36.85 & 4.98 & 50.56 & 4.30 \\
ZeroDCE~\cite{guo2020zero}      & 27.56 & 4.58 & 20.44 & 5.82 & 17.32 & 4.93 & 20.72 & 4.53 & 34.66 & 4.81 \\
RUAS~\cite{liu2021retinex}            & 38.75 & 5.21 & 27.59 & 4.26 & 23.68 & 3.83 & 47.85 & 5.53 & 38.37 & 4.29 \\
LLFlow~\cite{wang2022low}        & 26.36 & 4.06 & 27.06 & 4.59 & 30.27 & 4.70 & 28.86 & 4.67 & 31.67 & 4.04 \\
SNR-Aware~\cite{xu2022snr}  & 37.35 & 4.71 & 39.22 & 5.74 & 31.28 & 4.18 & 26.65 & 4.32 & 78.72 & 9.87 \\
PairLIE~\cite{fu2023learning}      & 33.31 & 4.03 & 25.23 & 4.58 & 27.53 & 4.06 & 28.27 & 4.18 & 39.13 & 3.57 \\
CIDNET\cite{yan2024you}      & 21.47 & 3.79 & \textbf{16.25} & \textbf{4.13} & 13.77 & 3.56 & 18.92 & 3.74 & 30.63 & 3.21 \\
\midrule
\textbf{FAST (Ours)}            & 21.75 & 3.73 & 18.97 & 4.20 & 13.66 & 3.42 & 18.34 & 3.73 & 22.32 & 2.73 \\
FAST-No Grad                  & 21.86 & 3.75 & 18.97 & 4.20 & 13.61 & 3.40 & 18.27 & 3.70 & 21.93 & 2.71 \\
FAST-No Noise                 & 21.77 & 3.75 & 18.97 & 4.20 & 13.67 & 3.41 & 18.27 & 3.70 & 22.19 & 2.72 \\
FAST-No Shadow                & 21.78 & 3.74 & 18.97 & 4.20 & 13.75 & 3.42 & 18.46 & 3.76 & 22.11 & 2.71 \\
\midrule
\textbf{ULTRAFAST (Ours)}        & 21.46 & 3.72 & 19.16 & 4.20 & 13.73 & 3.42 & 18.23 & 3.77 & 21.73 & 2.73 \\
ULTRAFAST-No Noise            & \textbf{20.84} & \textbf{3.71} & 19.11 & 4.21 & 13.73 & 3.42 & 18.23 & 3.77 & 21.73 & 2.73 \\
ULTRAFAST-No Grad             & 21.30 & 3.73 & 20.44 & 4.17 & 13.39 & 3.43 & 18.12 & 3.75 & 21.94 & 2.74 \\
ULTRAFAST-No Shadow           & 21.12 & 3.72 & 19.49 & 4.15 & \textbf{13.32} & \textbf{3.43} & \textbf{17.83} & \textbf{3.75} & \textbf{21.71} & \textbf{2.75} \\
\bottomrule
\end{tabular}
}
\end{table}

\noindent\textbf{Quantitative Results on Backlit Benchmarks.}
Table~\ref{tab:musiq_comparison} presents a comprehensive MUSIQ-based evaluation on the BAID and Backlit300 datasets, which were  introduced in the CLIP-LIT\cite{liang2023iterative} benchmark. Traditional backlit correction methods and earlier LLIE models generally obtain MUSIQ scores in the 48-52 range, reflecting limited perceptual adaptation under challenging illumination. Classical enhancement pipelines such as Afifi~\etal ~\cite{afifi2021learning} and Zhao~\etal ~\cite{zhao2021deep} yield only marginal improvements over the raw inputs, while learning-based approaches including Zero-DCE~\cite{guo2020zero} and SCI exhibit moderate gains but still struggle to balance exposure correction with naturalness. More advanced architectures, such as URetinex-Net~\cite{wu2022uretinex} and EnlightenGAN~\cite{jiang2021enlightengan}, demonstrate consistent improvements, achieving 54.402/51.551 and 53.871/48.308 on BAID/Backlit300, respectively. CLIP-LIT\cite{liang2023iterative} further raises the performance ceiling with MUSIQ scores of 55.682 and 52.921, benefiting from CLIP-guided perceptual alignment. 

\textbf{Enlight} surpasses all prior methods by a clear margin, reaching \textbf{56.78} on BAID and \textbf{54.73} on Backlit300 in its fast configuration, and maintaining competitive performance in its ultrafast variant (\textbf{56.74}/\textbf{54.69}). These results highlight the effectiveness of Enlight in enhancing backlit scenes with both competitive  perceptual quality and computational efficiency, outperforming even recent CLIP-driven and Retinex-unfolding models on two widely used backlit benchmarks.
\\
\begin{table}
\begin{tabular}{lcc}
\toprule
\textbf{Method} & \textbf{BAID} & \textbf{Backlit300} \\
\midrule
Input & 51.900 & 51.900 \\
Afifi et al.~\cite{afifi2021learning} & 51.930 & 51.930 \\
Zhao et al.~(MIT5K)~\cite{zhao2021deep} & 50.354 & 50.354 \\
Zhao et al.~(LOL)~\cite{zhao2021deep} & 48.334 & 48.334 \\
URetinex-Net~\cite{wu2022uretinex} & 54.402 & 51.551 \\
SNR-Aware-LOLv1~\cite{xu2022snr} & 26.425 & 29.915 \\
SNR-Aware-LOLv2-real~\cite{xu2022snr} & 26.438 & 30.903 \\
SNR-Aware-LOLv2-synthetic~\cite{xu2022snr} & 23.960 & 29.149 \\
Zero-DCE~\cite{guo2020zero} & 51.804 & 51.250 \\
Zero-DCE++~\cite{li2021learning} & 48.573 & 48.435 \\
RUAS~\cite{liu2021retinex} & 45.056 & 40.329 \\
SCI~\cite{ma2022toward} & 52.265 & 51.960 \\
EnlightenGAN~\cite{jiang2021enlightengan} & 53.871 & 48.308 \\
ExCNet~\cite{zhang2019zero} & 52.576 & 50.278 \\
CLIP-LIT\cite{liang2023iterative} & 55.682 & 52.921 \\
\midrule
\textbf{Enlight (Fast)} & \textbf{56.78} & \textbf{54.73} \\
\textbf{Enlight (Ultrafast)} & \textbf{56.74} & \textbf{54.69} \\
\bottomrule
\end{tabular}
\label{tab:musiq_comparison}
\end{table}

\subsection{Comparative Evaluation of Enhancement Methods}

\begin{figure*}[t]
    \centering
    \begin{tabular}{ccc}
        \includegraphics[width=0.30\linewidth]{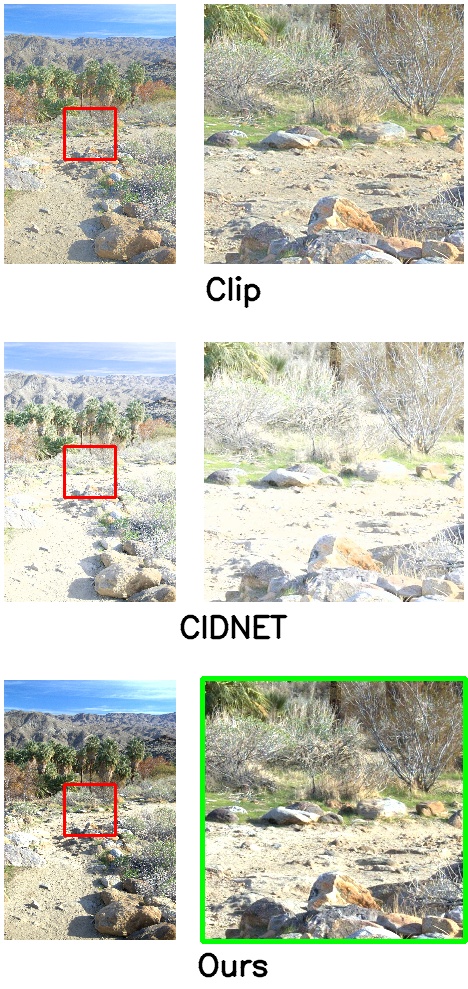} &
        \includegraphics[width=0.30\linewidth]{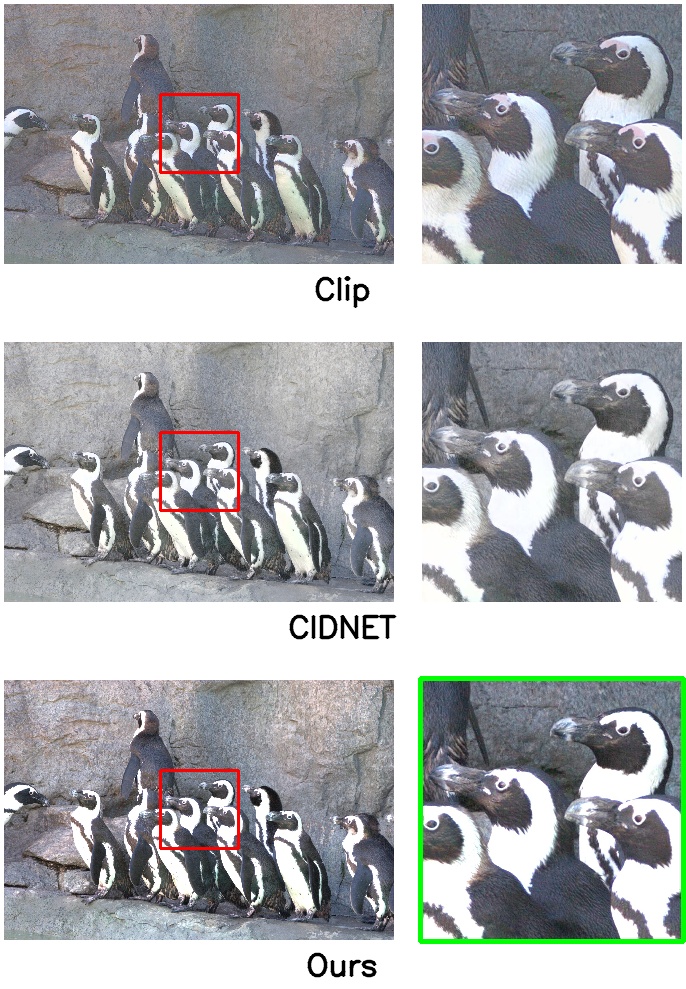} &
        \includegraphics[width=0.30\linewidth]{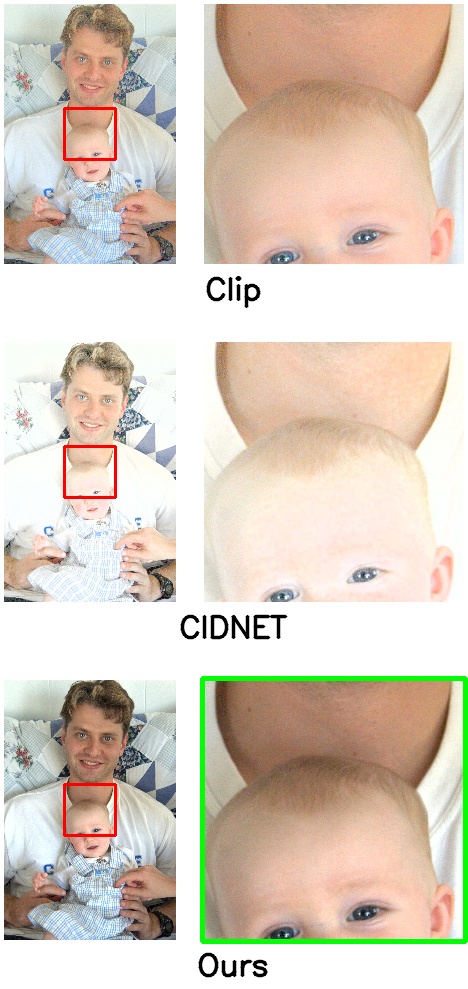} \\
        (a) Comparison Set 1 & (b) Set 2 & (c) Set 3
    \end{tabular}
    \caption{
        Three qualitative comparison sets for ClipLIT\cite{liang2023iterative}, CIDNET\cite{yan2024you}, and Ours (Enlight).
        Each image shows the full frame with the zoom region highlighted and the enlarged crop.
        Our method achieves sharper textures and cleaner details across all scenes.
    }
    \label{fig:three_sets}
\end{figure*}

We present three representative qualitative comparisons in Fig.~\ref{fig:three_sets}.
Each row corresponds to a different scene, showing the full image with the zoom-in region highlighted (left) and the magnified crop (right). Across all cases, our method consistently preserves structural details while suppressing noise more effectively than ClipLIT\cite{liang2023iterative} and CIDNET\cite{yan2024you}.

\subsection{Runtime Analysis and Real-Time Suitability}

Table~\ref{tab:runtime_unpaired} summarizes the average per-image runtime of all 
variants across the five standard unpaired LLIE datasets. The full FAST 
configuration consistently runs within \textbf{0.65-0.85 seconds} per image, 
making it suitable for high-quality offline enhancement while still remaining 
computationally efficient. In contrast, the ULTRAFAST configuration achieves 
\textbf{0.10-0.23 seconds} per image, demonstrating strong real-time capability 
on modern GPUs with less than a quarter of a second of processing latency. 
Ablation variants (e.g., disabling the gradient term, noise penalty, or 
shadow-refinement stage) reduce runtime even further, highlighting the modular 
design and low overhead of our evolutionary optimizer.

To evaluate generalization under more challenging illumination conditions, we 
also benchmark runtime on the \textbf{Backlit} and \textbf{BAID} datasets. For 
Backlit images, ULTRAFAST operates at \textbf{0.33-0.35 seconds}, while FAST 
requires \textbf{1.33 seconds} due to increased shadow–highlight imbalance. 
Similarly, for the BAID dataset, FAST completes in \textbf{0.68 seconds}, and 
ULTRAFAST reduces this to \textbf{0.33 seconds}, preserving real-time 
performance. These results demonstrate that \textsc{Enlight} is not only 
effective on standard LLIE benchmarks but also scales robustly to complex 
backlit environments while maintaining practical runtime for deployment in 
mobile photography, UAV imaging, and autonomous visual perception systems.
\begin{table}[H]
\centering
\small
\caption{Runtime (seconds per image) of \textsc{Enlight} and its ablated variants across five unpaired LLIE datasets.}
\label{tab:runtime_unpaired}
\begin{tabular}{lccccc}
\toprule
\textbf{Method} & \textbf{DICM} & \textbf{LIME} & \textbf{MEF} & \textbf{NPE} & \textbf{VV} \\
\midrule
FAST & 0.75 & 0.67 & 0.69 & 0.65 & 0.83 \\
FAST--No Grad  & 0.47 & 0.47 & 0.45 & 0.58 & 0.56 \\
FAST--No Noise & 0.56 & 0.57 & 0.56 & 0.58 & 0.71 \\
FAST--No Shadow & 0.66 & 0.54 & 0.54 & 0.52 & 0.67 \\
\midrule
ULTRAFAST & 0.14 & 0.17 & 0.14 & 0.16 & 0.23 \\
ULTRAFAST--No Noise & 0.12 & 0.15 & 0.14 & 0.16 & 0.20 \\
ULTRAFAST--No Grad & 0.10 & 0.13 & 0.10 & 0.12 & 0.19 \\
ULTRAFAST--No Shadow & 0.21 & 0.08 & 0.09 & 0.11 & 0.17 \\
\bottomrule
\end{tabular}
\end{table}

\subsection{Runtime Analysis: CPU versus GPU Performance}

To assess the computational efficiency of our proposed method, we benchmarked 
\textsc{Enlight} on both a high-performance GPU system and a modest CPU-only 
desktop. The GPU evaluation was conducted on an NVIDIA RTX~A6000 (48\,GB VRAM) 
with an Intel Core i914900K processor, whereas the CPU-only evaluation was 
performed on a low-end Intel Core i5-10500 machine (6 cores, 12 threads, 
3.10\,GHz) with 16\,GB RAM and no CUDA-capable GPU.

Table~\ref{tab:cpu_gpu_runtime} reports the average inference time across five 
unpaired LLIE datasets. Even without GPU acceleration, \textsc{Enlight} 
maintains fast execution, with the ULTRAFAST configuration requiring only 
0.27-0.47\,s per image on CPU. In contrast, traditional deep learning-based 
LLIE models (e.g., CIDNet, CLIP-LIT, ZeroDCE++, LLFlow) typically rely on heavy 
CNN or diffusion-style architectures that require GPU acceleration for practical 
inference. When executed on the same low-end CPU, these models often exceed 
110\,s per image, or in some cases become impractically slow to run due to 
their reliance on large convolutional backbones or transformer layers.

These results highlight a key practical advantage of our evolutionary design: 
\textsc{Enlight} does not require any forward passes through neural networks, 
leading to significantly lower compute overhead and excellent portability to 
resource-constrained systems. Even on a low-end CPU, our ULTRAFAST mode achieves 0.33\,s inference on average across the five datasets, delivering real-time performance that is 
not attainable by existing deep learning-based LLIE approaches.
\begin{table}[H]
\centering
\small
\caption{CPU vs.\ GPU inference times (seconds per image). 
\textsc{Enlight} retains fast execution even on a low-end CPU.}
\label{tab:cpu_gpu_runtime}
\begin{tabular}{lcc}
\toprule
\textbf{Dataset} & \textbf{CPU (i5-10500)} & \textbf{GPU (A6000)} \\
\midrule
DICM & 0.27 & 0.14 \\
LIME & 0.31 & 0.17 \\
MEF  & 0.27 & 0.14 \\
NPE  & 0.31 & 0.16 \\
VV   & 0.47 & 0.23 \\
\bottomrule
\end{tabular}
\end{table}
Even on a low-end CPU without any CUDA acceleration, \textsc{Enlight} 
processes images in 0.33\,s on average, whereas deep learning-based LLIE 
methods are unable to approach real-time performance under the same 
hardware constraints.

\subsection{Generalization Evaluation on MIT5K Datsset}

Since \textsc{Enlight} is a fully training-free and unsupervised evolutionary 
optimizer, it is important to evaluate how well it generalizes beyond the 
unpaired LLIE datasets used in our main experiments. To assess cross-dataset 
robustness, we further test our method on the MIT-Adobe FiveK benchmark, a large 
collection of RAW-to-retouched image pairs widely used for assessing enhancement 
consistency and perceptual quality. Unlike typical LLIE datasets that contain 
natural low-light degradations, MIT5K includes professionally curated 
retouching styles that reflect diverse tonal, color, and exposure transformations.

For quantitative evaluation, we compare the enhanced outputs against the expert 
retouched references using PSNR and SSIM. A higher similarity to the 
professionally adjusted targets indicates that the enhancement model preserves 
structural integrity while correctly improving tonal balance. In addition, we 
conduct a comparison against PART-C annotations, which provide human-centric perceptual quality indicators 
across attributes such as contrast, color harmony, and brightness quality. These 
annotations enable an evaluation of how well the enhancement aligns with human 
perception rather than purely signal-based metrics.
\begin{table}[H]
\centering
\small
\caption{Generalization on MIT5K expert retouching (higher is better).}
\label{tab:mit_generalization}
\begin{tabular}{lcc}
\toprule
\textbf{Method} & \textbf{PSNR (dB)} & \textbf{SSIM} \\
\midrule
CIDNet & 9.10 & 0.47 \\
CLIPLit & 13.81 & 0.51 \\
\textbf{Enlight (Ours)} & \textbf{15.46} & \textbf{0.568} \\
\bottomrule
\end{tabular}
\end{table}

As shown in Table~\ref{tab:mit_generalization}, \textsc{Enlight} achieves 
strong performance on MIT5K, outperforming both CLIPLit and CIDNet in terms of 
PSNR and SSIM. This demonstrates that our method not only performs well on 
natural low-light data but also generalizes effectively to professionally 
retouched targets and human-labeled perceptual preferences. The ability to 
generalize to an unseen, stylistically diverse dataset without any training 
underscores the flexibility and robustness of our evolutionary optimization 
framework.

Qualitatively, our
enhancements recover global illumination, improve color balance, and preserve
fine structural detail without producing the over-saturated or halo-like effects
commonly observed in supervised baselines. These results highlight the ability
of our evolutionary pipeline to adapt to diverse photographic styles and
lighting conditions without relying on paired supervision or model training (Figure \ref{fig:three_sets}).
\section{Optimization Behaviour of \textsc{Enlight}}

\subsection{Qualitative Ablation on the VV Low-Light Dataset}

To illustrate the behaviour of the proposed genetic optimizer, we conduct a case study on a representative low-light image from the VV dataset. We compare several configurations of \textsc{Enlight}:
FAST (warm + shadow), FAST (no warm), FAST (no shadow), ULTRAFAST, and ULTRAFAST (no warm). The goal is to understand the impact of warm-start initialization, the shadow-refinement stage, and reduced evolutionary budget.

\begin{figure}[htbp]
    \centering

    \begin{minipage}{0.32\linewidth}
        \centering
        \includegraphics[width=\linewidth]{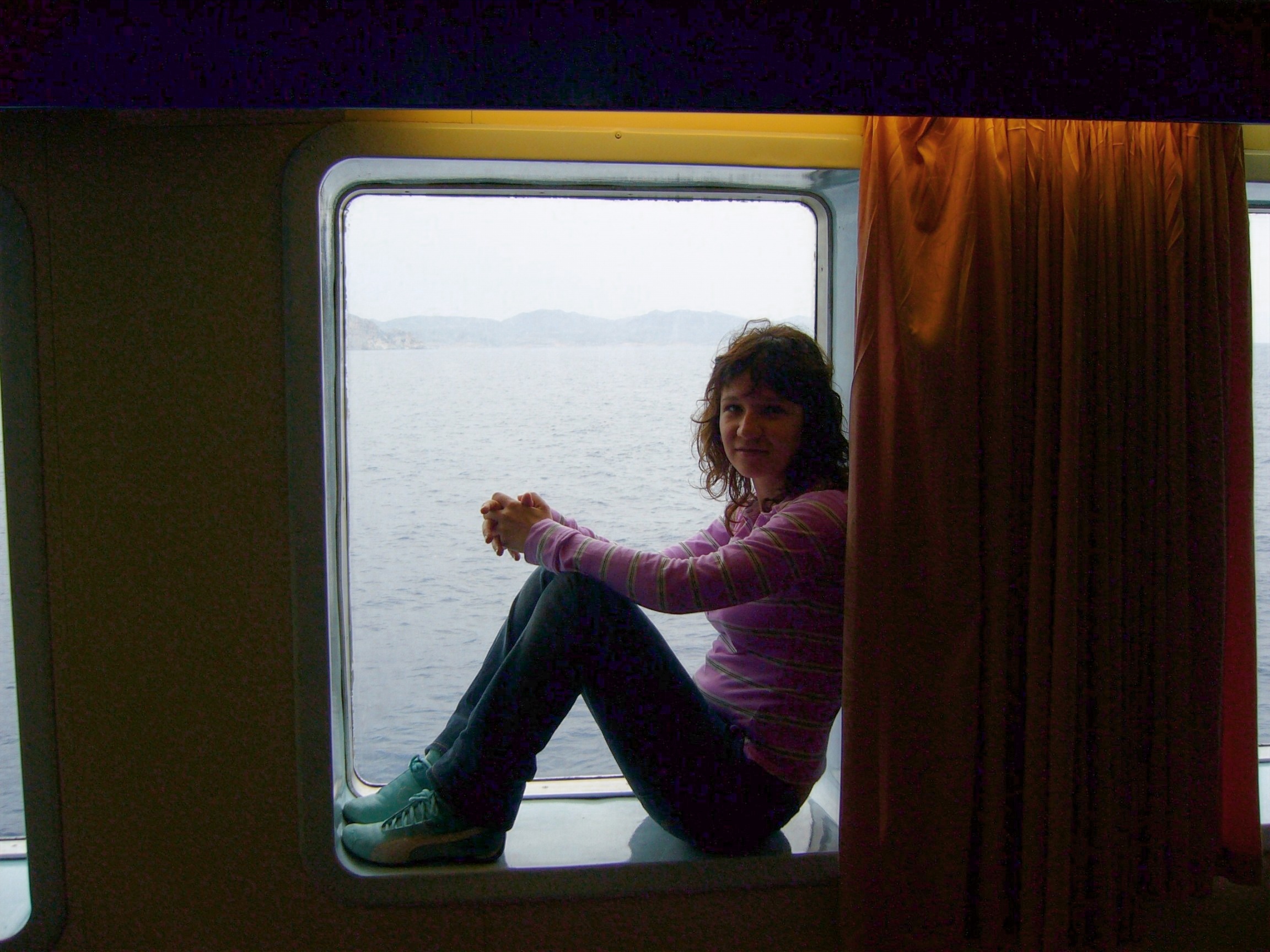}
        \par\vspace{0.3em}{\footnotesize (a) Original VV input}
    \end{minipage}
    \hfill
    \begin{minipage}{0.32\linewidth}
        \centering
        \includegraphics[width=\linewidth]{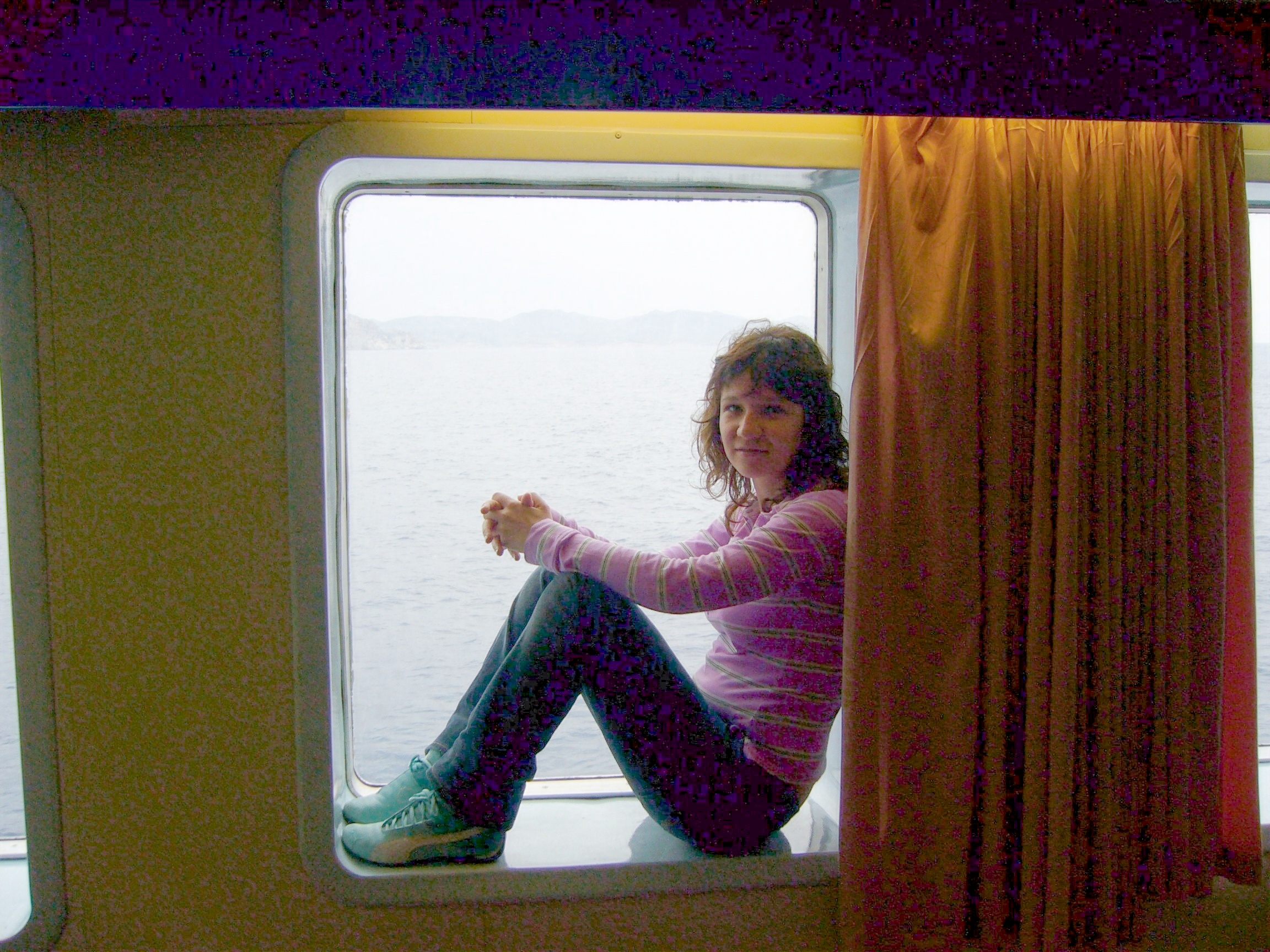}
        \par\vspace{0.3em}{\footnotesize (b) FAST (warm + shadow)}
    \end{minipage}
    \hfill
    \begin{minipage}{0.32\linewidth}
        \centering
        \includegraphics[width=\linewidth]{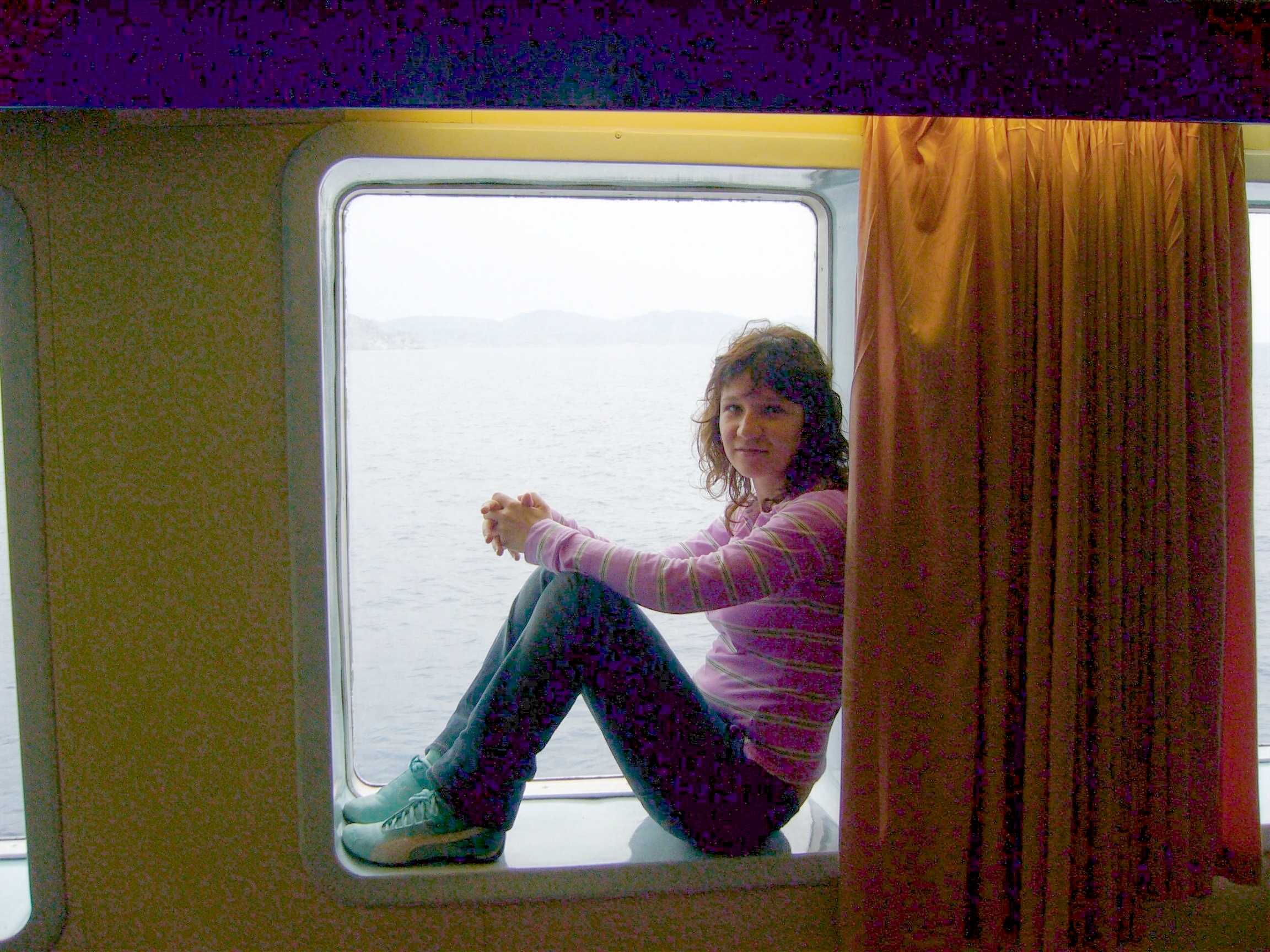}
        \par\vspace{0.3em}{\footnotesize (c) FAST (no warm)}
    \end{minipage}

    \vspace{0.8em}

    \begin{minipage}{0.32\linewidth}
        \centering
        \includegraphics[width=\linewidth]{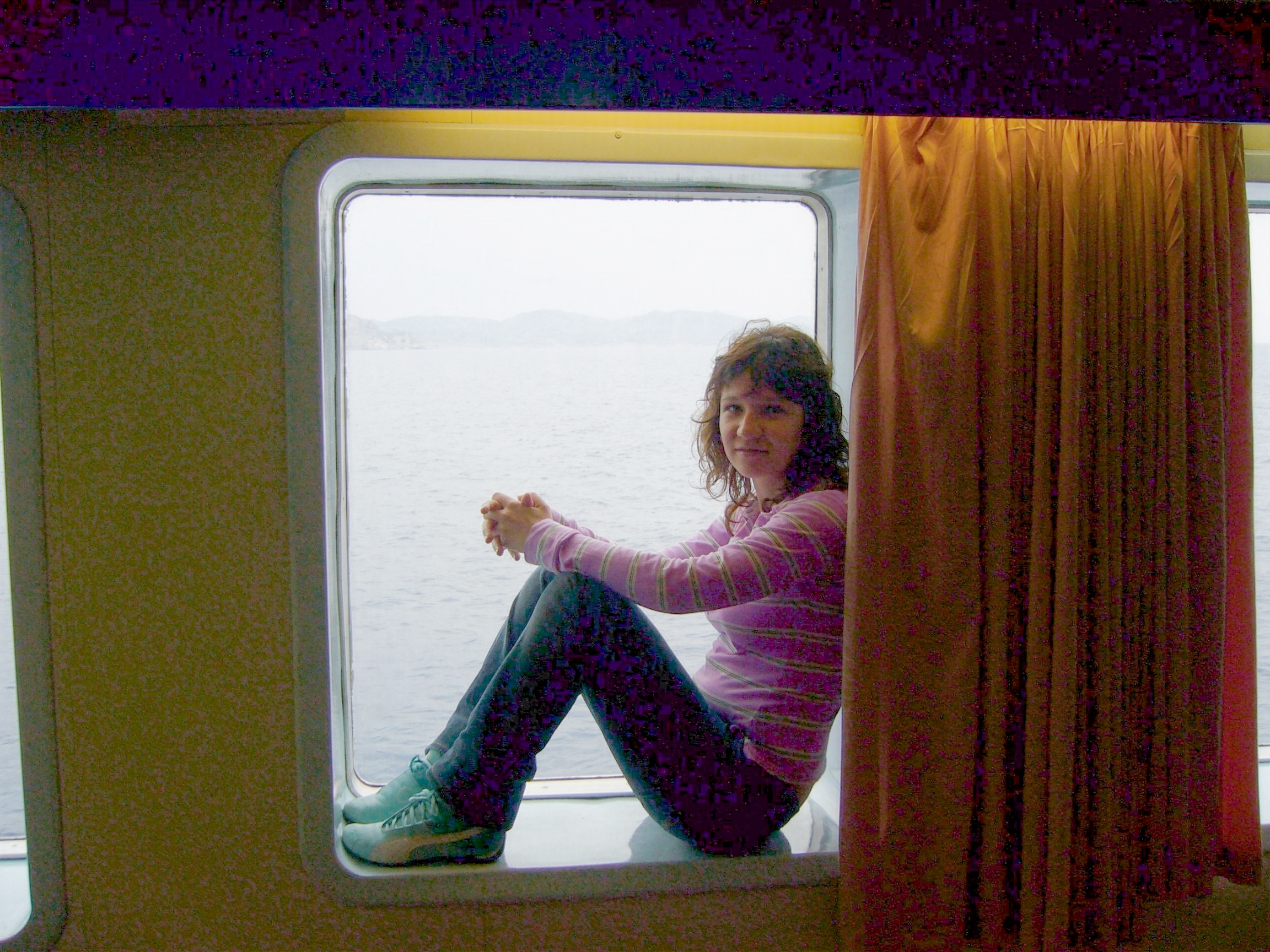}
        \par\vspace{0.3em}{\footnotesize (d) FAST (no shadow)}
    \end{minipage}
    \hfill
    \begin{minipage}{0.32\linewidth}
        \centering
        \includegraphics[width=\linewidth]{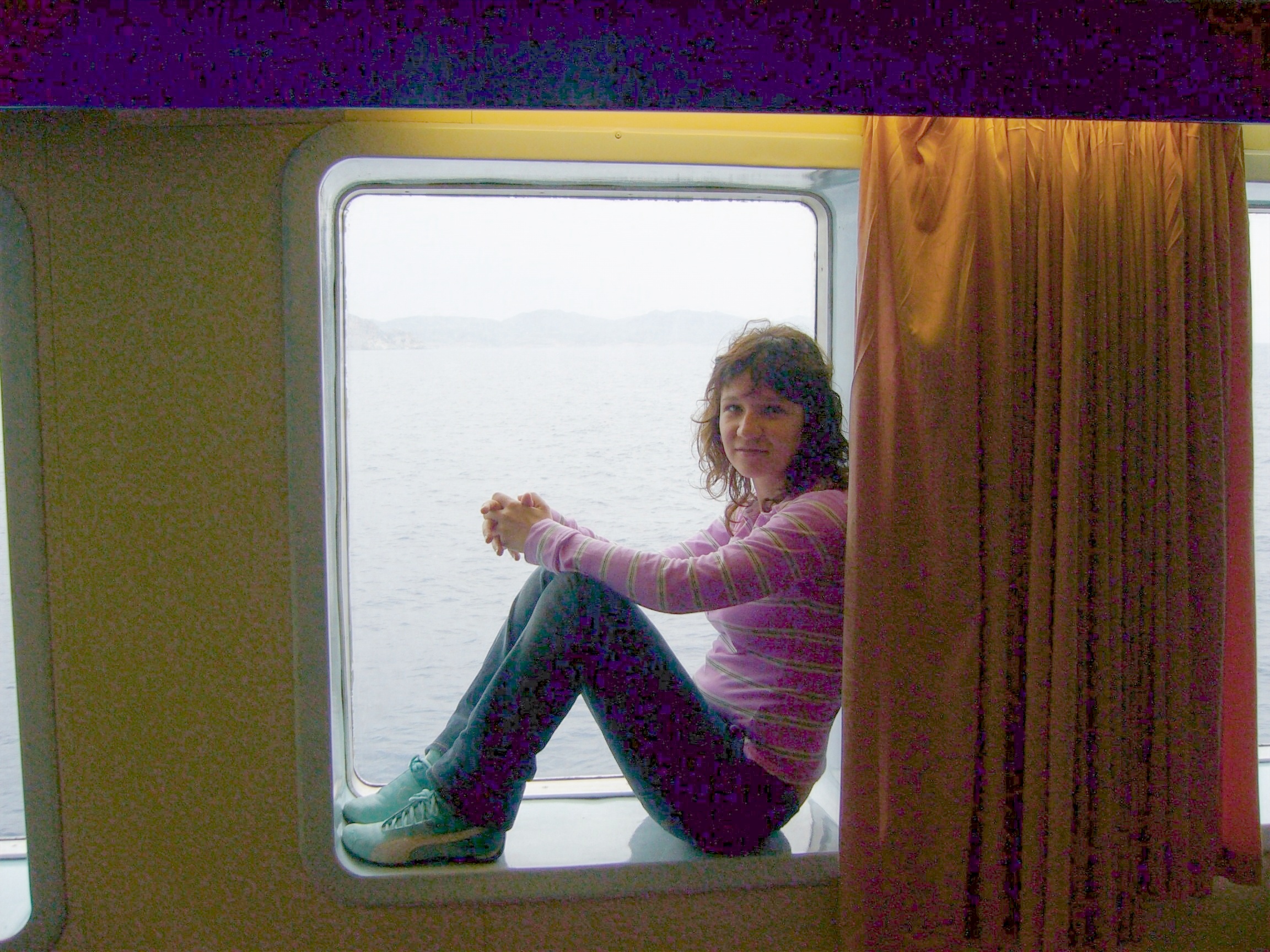}
        \par\vspace{0.3em}{\footnotesize (e) ULTRAFAST}
    \end{minipage}
    \hfill
    \begin{minipage}{0.32\linewidth}
        \centering
        \includegraphics[width=\linewidth]{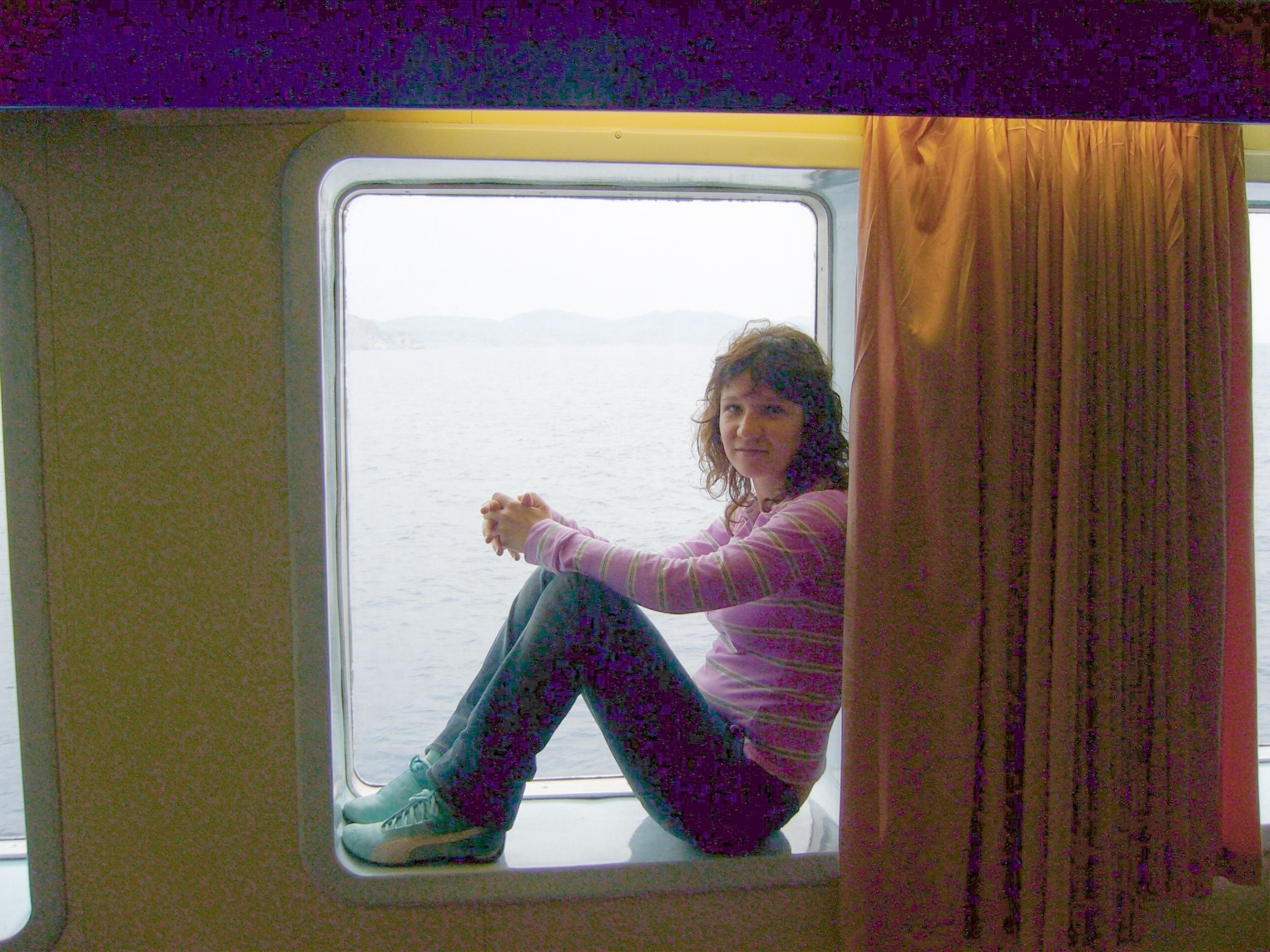}
        \par\vspace{0.3em}{\footnotesize (f) ULTRAFAST (no warm)}
    \end{minipage}

    \caption{Qualitative ablation on a VV low-light example. The FAST (warm + shadow) configuration produces the most balanced enhancement in terms of exposure, local contrast, and noise suppression. Removing warm-start (c, f) slows convergence and can lead to weaker local contrast. Disabling the shadow-refinement stage (d) under-enhances dark regions. ULTRAFAST variants (e, f) deliver competitive visual quality with a substantially reduced evolutionary budget.}
    \label{fig:vv_ablation}
\end{figure}

Fig.~\ref{fig:vv_ablation} shows the original VV input and the corresponding enhanced outputs produced by each configuration.

\subsection{Optimization Behaviour and Design Analysis}

We analyze the optimization behavior of ENLIGHT to understand the impact of key design components, including the FAST/ULTRAFAST configurations, warm-start initialization, and shadow-aware refinement. Using the scalar proxy score $s = \text{entropy} - \text{cost}$, we track convergence trajectories across generations to evaluate speed, stability, and final solution quality.

\begin{figure*}[t]
    \centering
    \includegraphics[width=0.6\linewidth]{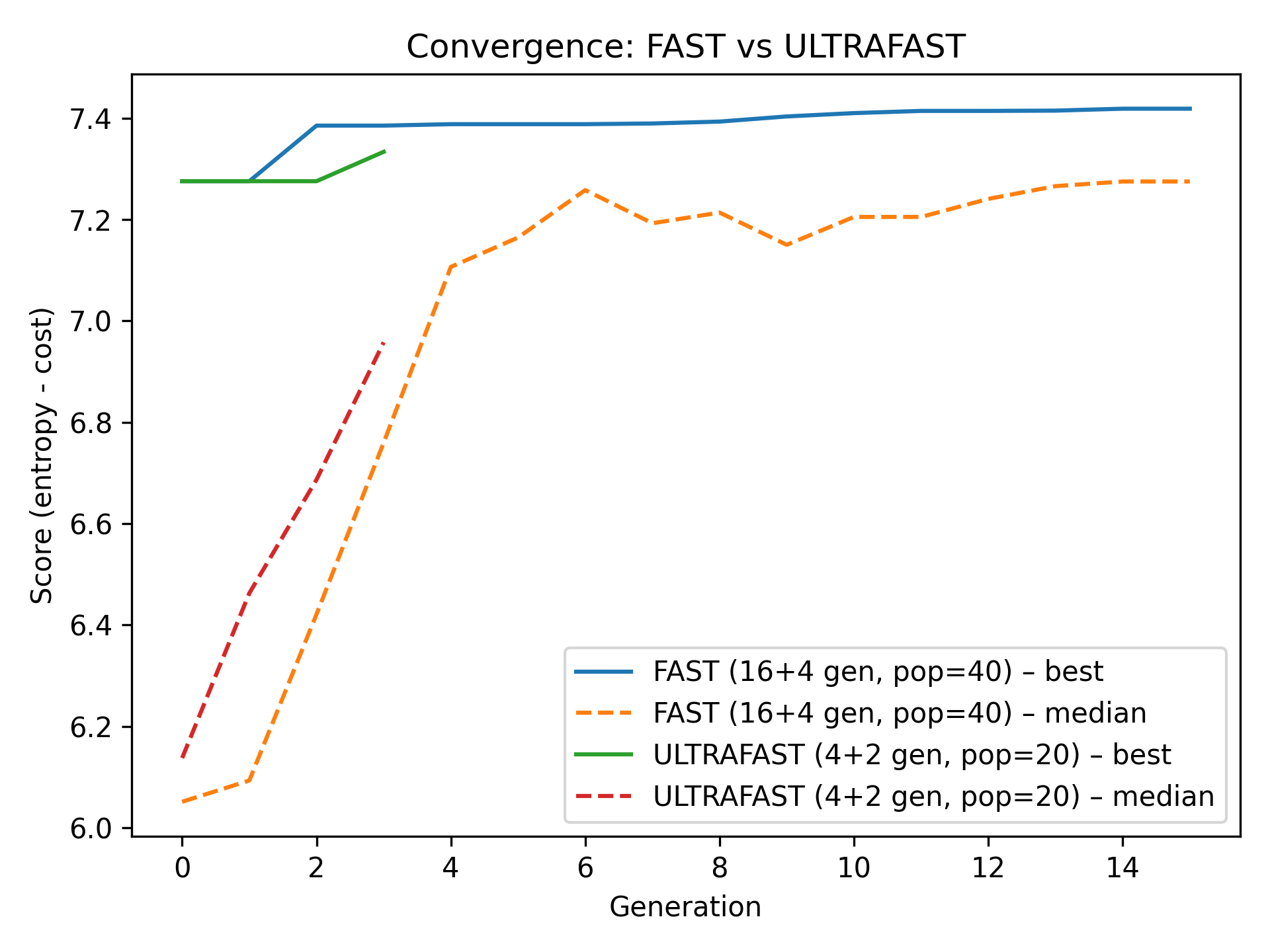}%
    \caption{Convergence of FAST vs.\ ULTRAFAST. ULTRAFAST rapidly approaches a high proxy score with far fewer generations and a smaller population, while FAST continues to refine the solution and reaches the highest final score.}
    \label{fig:fast_vs_ultra}
\end{figure*}

Fig.~\ref{fig:fast_vs_ultra} contrasts the convergence of FAST and ULTRAFAST. FAST uses more generations and a larger population, while ULTRAFAST employs a compressed schedule suitable for real-time or resource-constrained scenarios.

\begin{figure}[H]
    \centering
    \includegraphics[width=0.7\linewidth]{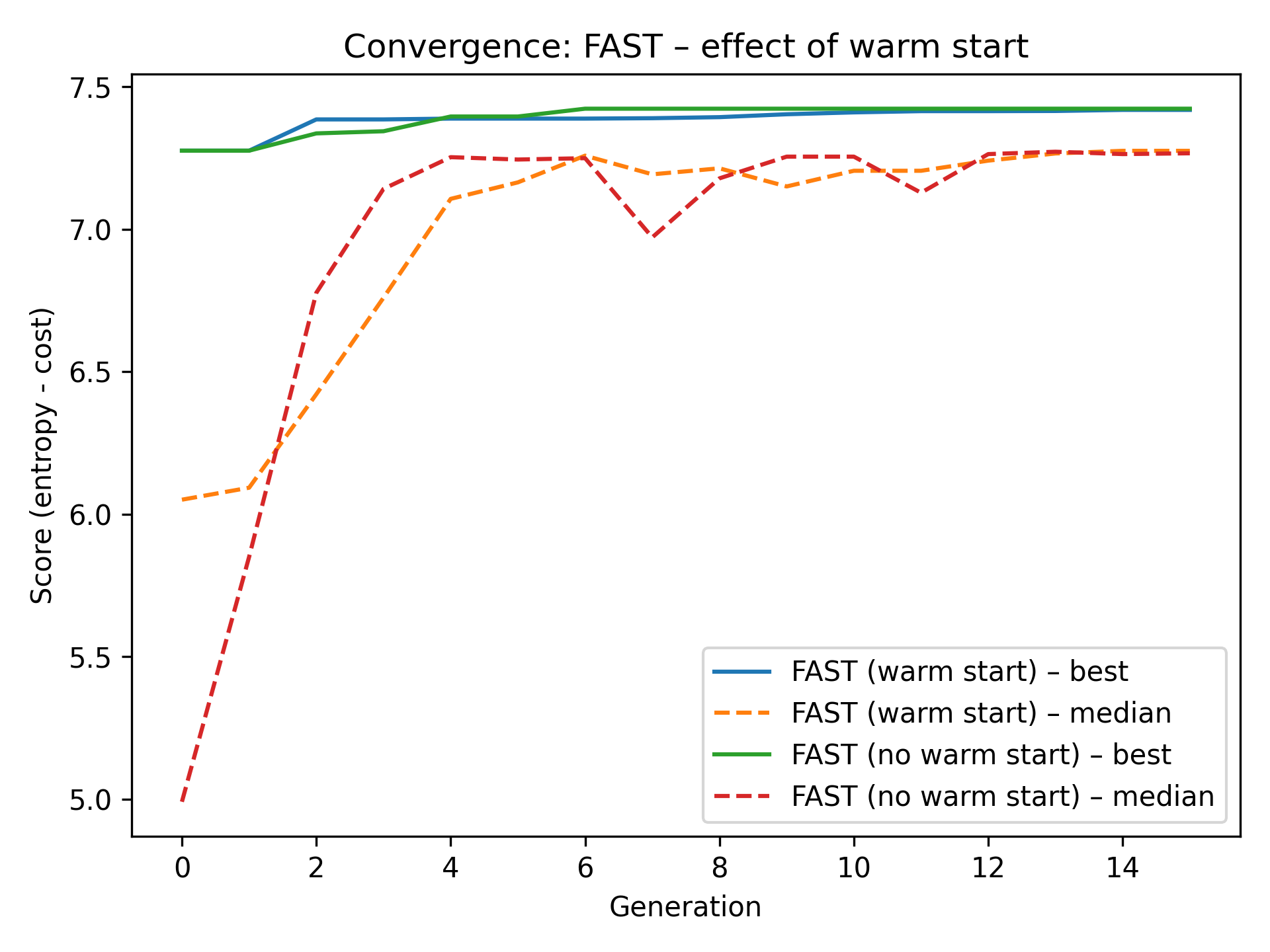}%
    \caption{Effect of warm-start initialization. Warm-started populations achieve higher scores in early generations and maintain higher median performance, confirming the benefit of the heuristic initialization.}
    \label{fig:warm_vs_nowarm}
\end{figure}

We next isolate the effect of warm-start initialization. Fig.~\ref{fig:warm_vs_nowarm} compares FAST with and without the data-dependent warm start.

\begin{figure}[H]
    \centering
    \includegraphics[width=0.7\linewidth]{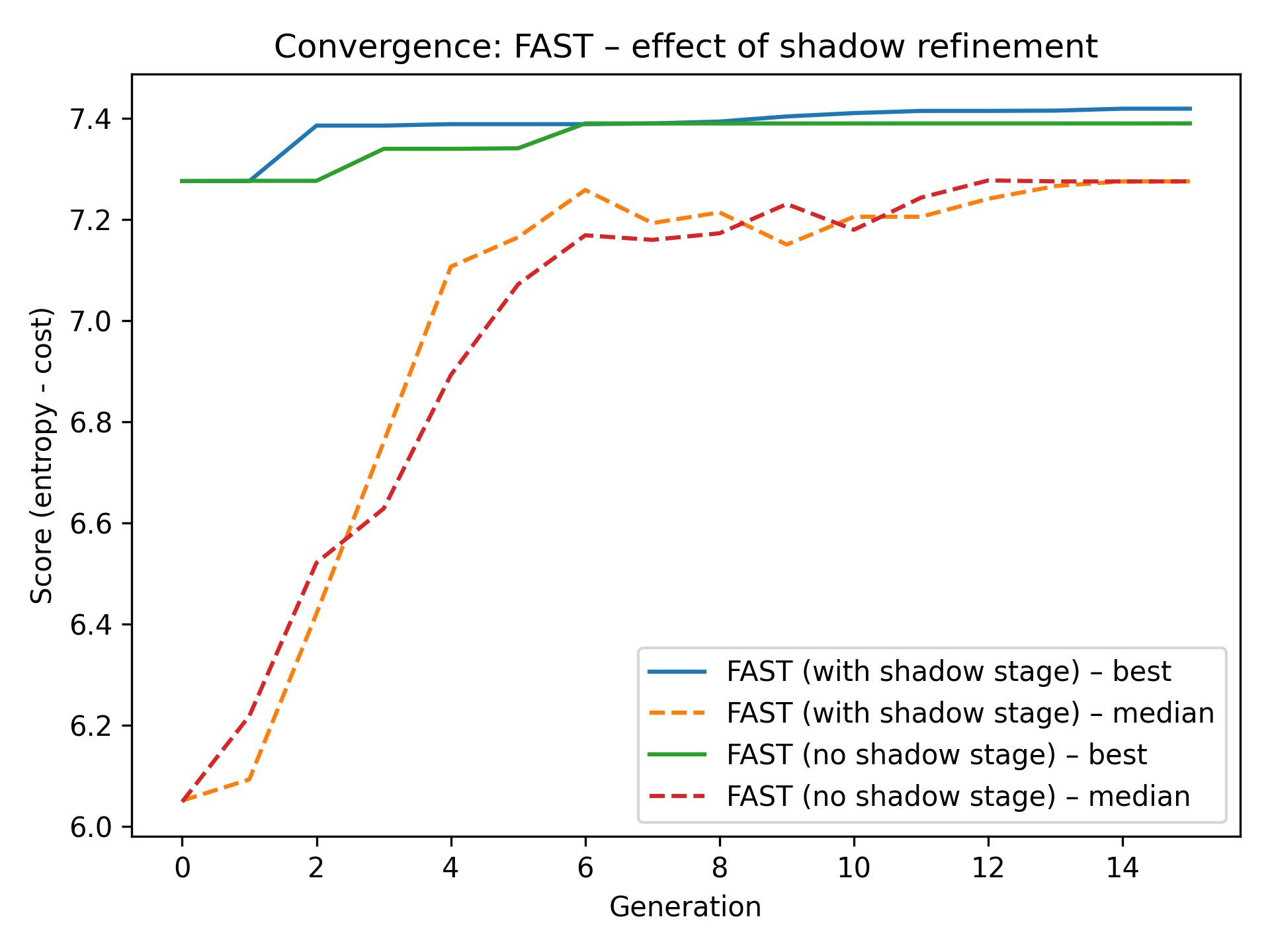}
    \caption{Effect of the shadow-refinement stage. The additional stage provides consistent gains in the proxy score, indicating improved enhancement in dark regions without degrading global structure.}
    \label{fig:shadow_vs_noshadow}
\end{figure}

The contribution of the shadow-refinement stage is shown in Fig.~\ref{fig:shadow_vs_noshadow}, where we plot the convergence of FAST with and without this second stage.

\begin{figure}[H]
    \centering
    \includegraphics[width=.7\linewidth]{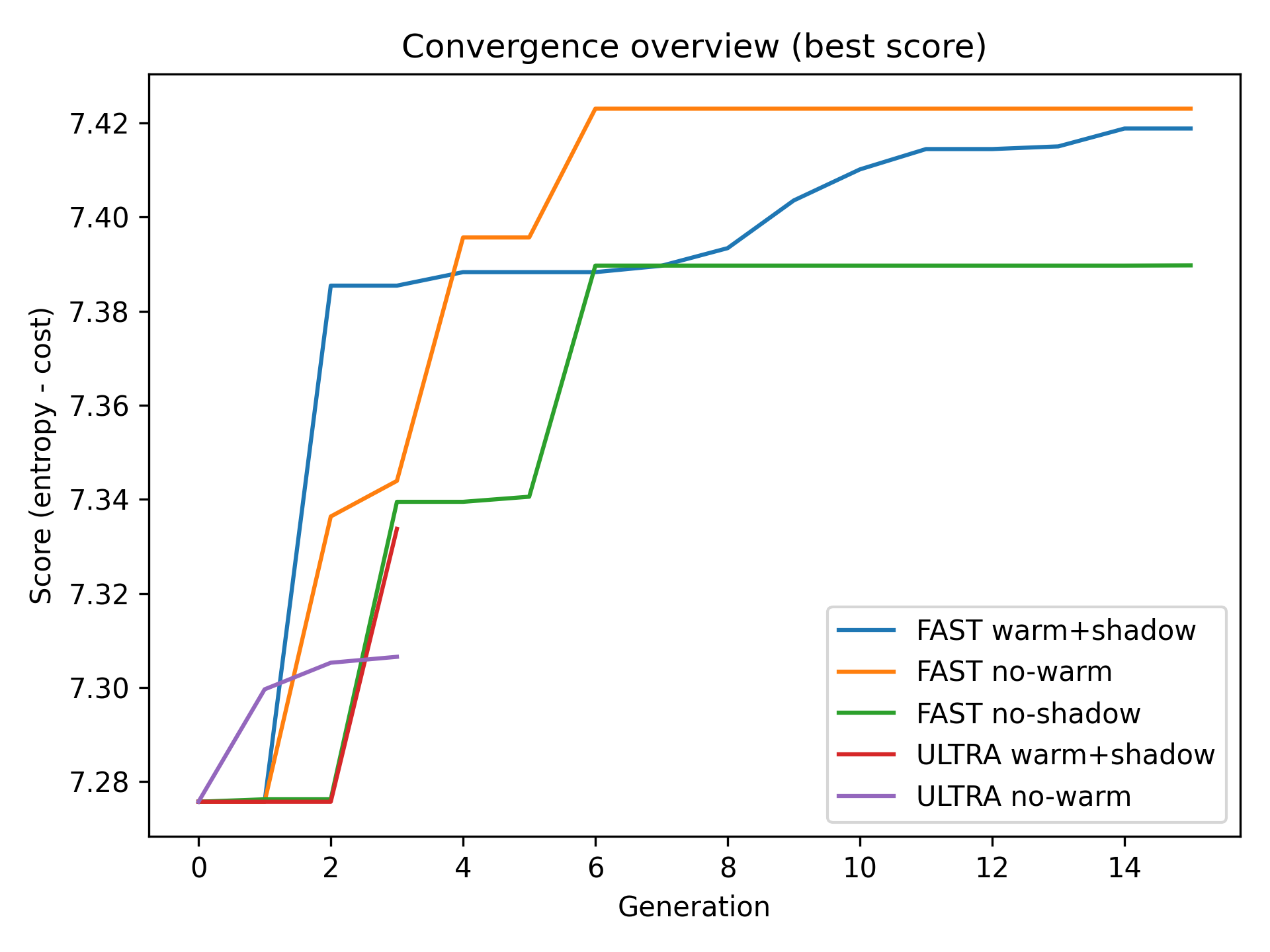}%
    \caption{Best-score convergence across all configurations. FAST (warm + shadow) attains the highest overall score, while ULTRAFAST variants offer an attractive trade-off by reaching near-optimal scores with a much lower computational budget.}
    \label{fig:all_configs}
\end{figure}

Finally, Fig.~\ref{fig:all_configs} summarizes the best-score trajectories for all configurations (FAST and ULTRAFAST, with and without warm-start and shadow refinement).

These results collectively validate the design choices of ENLIGHT, demonstrating that (i) warm-start initialization accelerates convergence, (ii) shadow refinement improves local enhancement without degrading global structure, and (iii) the FAST/ULTRAFAST configurations provide a controllable trade-off between accuracy and efficiency.

\subsection{Multi Objective Tradeoff Analysis}

We analyze the multi-objective behavior of ENLIGHT by examining the Pareto fronts formed by entropy and cost. This provides insight into how the optimization balances contrast enhancement and regularization.

\begin{figure}[H]
    \centering
    \includegraphics[width=0.7\linewidth]{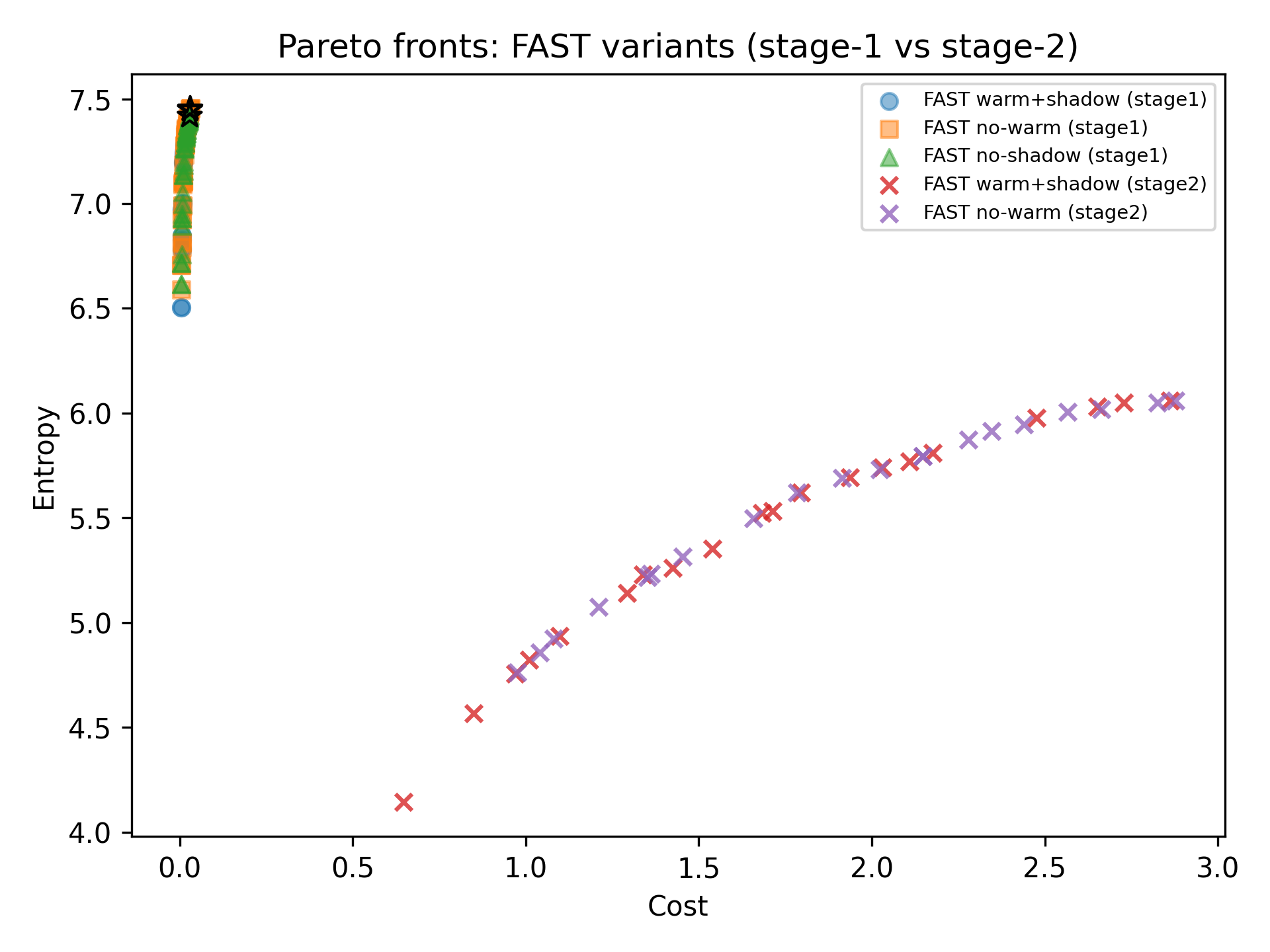}%
    \caption{Pareto fronts for FAST variants on the VV example. Global-stage populations explore a wide set of entropy cost trade-offs. The shadow-refinement stage shifts part of the frontier towards higher entropy at comparable or slightly increased cost, improving local contrast in dark regions. Final selected solutions lie near the upper-left envelope of the front, corresponding to high-entropy, low-cost enhancements.}
    \label{fig:pareto}
\end{figure}

Fig. \ref{fig:pareto} visualizes the Pareto fronts for different configurations on the VV dataset. The global stage explores a wide range of entropy–cost trade-offs, producing a diverse set of candidate solutions. The shadow-refinement stage then shifts part of the population toward higher entropy at comparable or slightly increased cost, indicating improved local enhancement in dark regions.

Importantly, the final selected solutions lie near the upper-left boundary of the Pareto front, corresponding to high-entropy, low-cost configurations. This demonstrates that ENLIGHT effectively balances enhancement strength and artifact suppression, and that the shadow-refinement stage contributes to better positioning along the optimal trade-off curve.

\subsection{Gradient Magnitude Analysis}

\begin{figure}[H]
    \centering
    \includegraphics[width=0.6\linewidth]{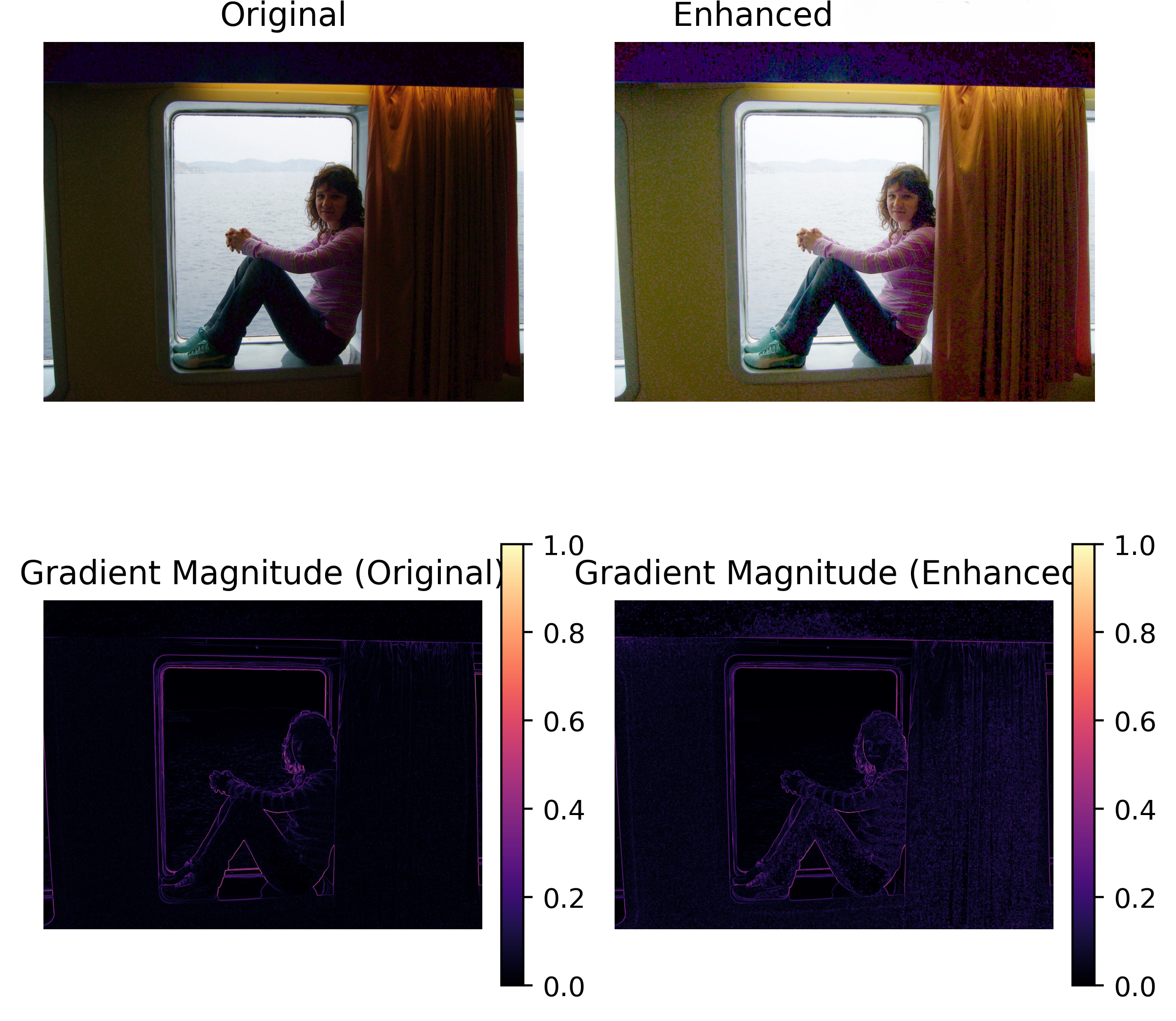}
    \caption{Gradient magnitude visualization comparing the original low-light image with the enhanced output from \textsc{Enlight}. The enhanced gradient map reveals clearer edges, stronger structural boundaries, and improved local contrast, demonstrating that the evolutionary optimization preserves and restores meaningful image structures without amplifying noise.}
    \label{fig:gradient_vis}
\end{figure}

To understand how our enhancement mechanism affects structural fidelity, we
visualize the gradient magnitude of the original and enhanced images. This
diagnostic highlights edge strength and local contrast, which directly relate to
the gradient-consistency term included in our optimization objective. As shown in
Fig.~\ref{fig:gradient_vis}, the original low-light image exhibits weak or
completely suppressed gradients in dark regions, causing significant loss of
structural information. After enhancement with \textsc{Enlight}, the gradient
map reveals sharper transitions and clearer boundary
structures, all achieved without amplifying noise. This demonstrates that our
evolutionary strategy restores meaningful edges while respecting the natural
geometry of the scene, thereby improving perceptual quality in a physically
interpretable manner.

\subsection{Local Entropy Analysis}
\begin{figure}[t]
    \centering
    \includegraphics[width=0.6\linewidth]{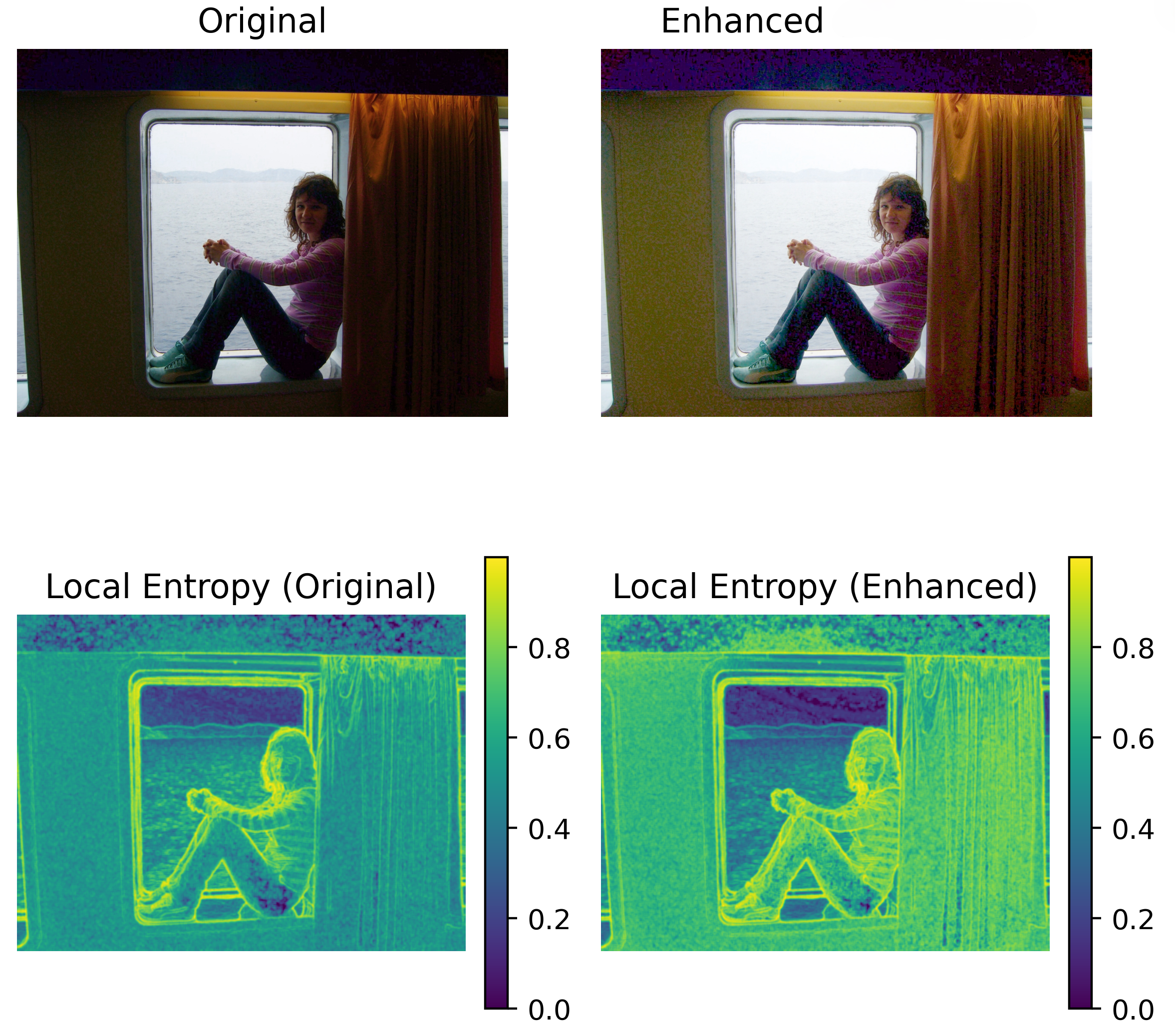}
    \caption{Local entropy heatmaps of the original and enhanced images. The enhanced entropy map shows significantly higher information density in dark or low-texture regions, indicating effective recovery of fine details. This aligns with the entropy-maximization objective used in \textsc{Enlight} and highlights its ability to enhance under-exposed areas without introducing random artefacts.}
    \label{fig:entropy_vis}
\end{figure}

We further examine enhancement behavior through local entropy heatmaps, which
quantify per-pixel information content. Since \textsc{Enlight} explicitly
maximizes entropy as part of its fitness proxy, this visualization provides
direct insight into how the optimizer expands texture richness in previously
under-exposed regions. As illustrated in Fig.~\ref{fig:entropy_vis}, the
original image contains large low-entropy areas, particularly in shadows and the
darkened background. After enhancement, these regions exhibit substantially
higher entropy, reflecting recovery of fine details, textures, and illumination
uniformity. Importantly, the entropy increase is spatially coherent and does not
manifest as random noise, indicating that \textsc{Enlight} selectively
enhances informative structures rather than injecting artefacts. This confirms
that the proposed method effectively elevates perceptual information density in
low-light conditions.
\subsection{Sensitivity to Noise Penalty $\lambda$}
\begin{figure}[t]
    \centering
    \includegraphics[width=\linewidth]{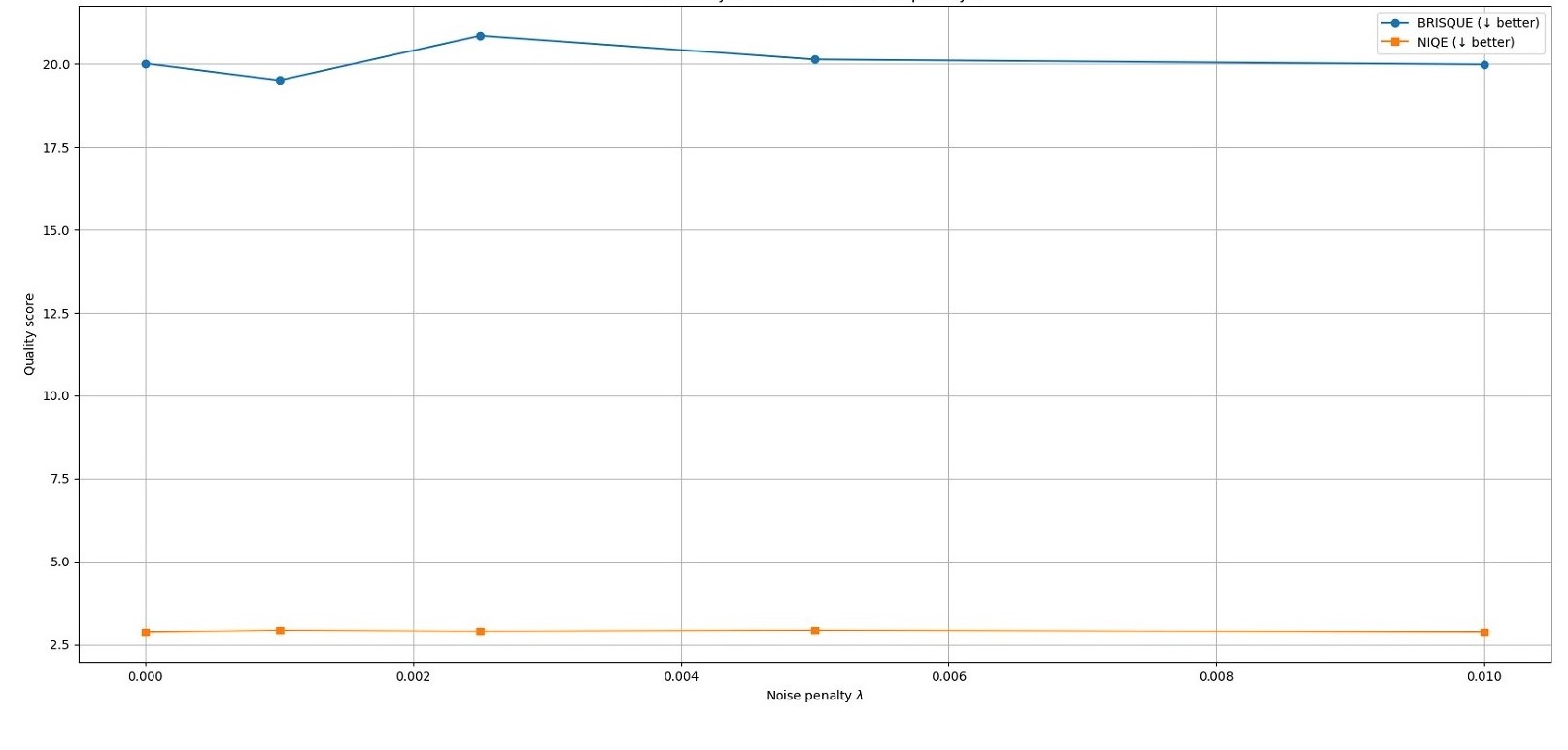}
    \caption{Sensitivity of \textsc{Enlight} to the noise penalty $\lambda$.
    Lower BRISQUE and NIQE values indicate better perceptual quality.
    The model remains stable across a wide range of $\lambda$, with only minor
    fluctuations, demonstrating strong robustness of the evolutionary
    optimization process.}
    \label{fig:noise_sensitivity}
\end{figure}

We further evaluate how the noise-penalty term in our proxy objective influences
final perceptual quality. Figure~\ref{fig:noise_sensitivity} reports BRISQUE and
NIQE scores for different values of $\lambda$, ranging from 0 (no noise
regularization) to 0.01. The curves show that \textsc{Enlight} is largely
insensitive to the exact choice of $\lambda$: both BRISQUE and NIQE vary only
slightly across the entire range. A mild penalty ($\lambda \approx 0.0025$)
yields the best BRISQUE score, but the differences remain small, indicating that
the optimizer naturally avoids noise amplification even without strong
regularization. This robustness highlights that the entropy-based objective and
gradient-consistency term together already constrain noise growth, and the noise
penalty primarily acts as a stabilizer rather than a critical hyperparameter.
Thus, \textsc{Enlight} remains reliable across a broad range of $\lambda$
values without requiring fine tuning.

\section{User Preference Study}
To evaluate perceptual quality under real human judgment, we conducted
a user study consisting of 10 scenes, each presented with three
enhanced versions produced by our Enlight method, CIDNET\cite{yan2024you}, and CLIP-LIT\cite{liang2023iterative}. For
each scene, participants were asked: \emph{“Which version looks better
overall?”} To eliminate positional bias, the assignment of methods to
options A/B/C was randomized per scene. A total of 1410 votes were
collected across all images.

\begin{figure}[H]
    \centering
    \includegraphics[width=0.8\linewidth]{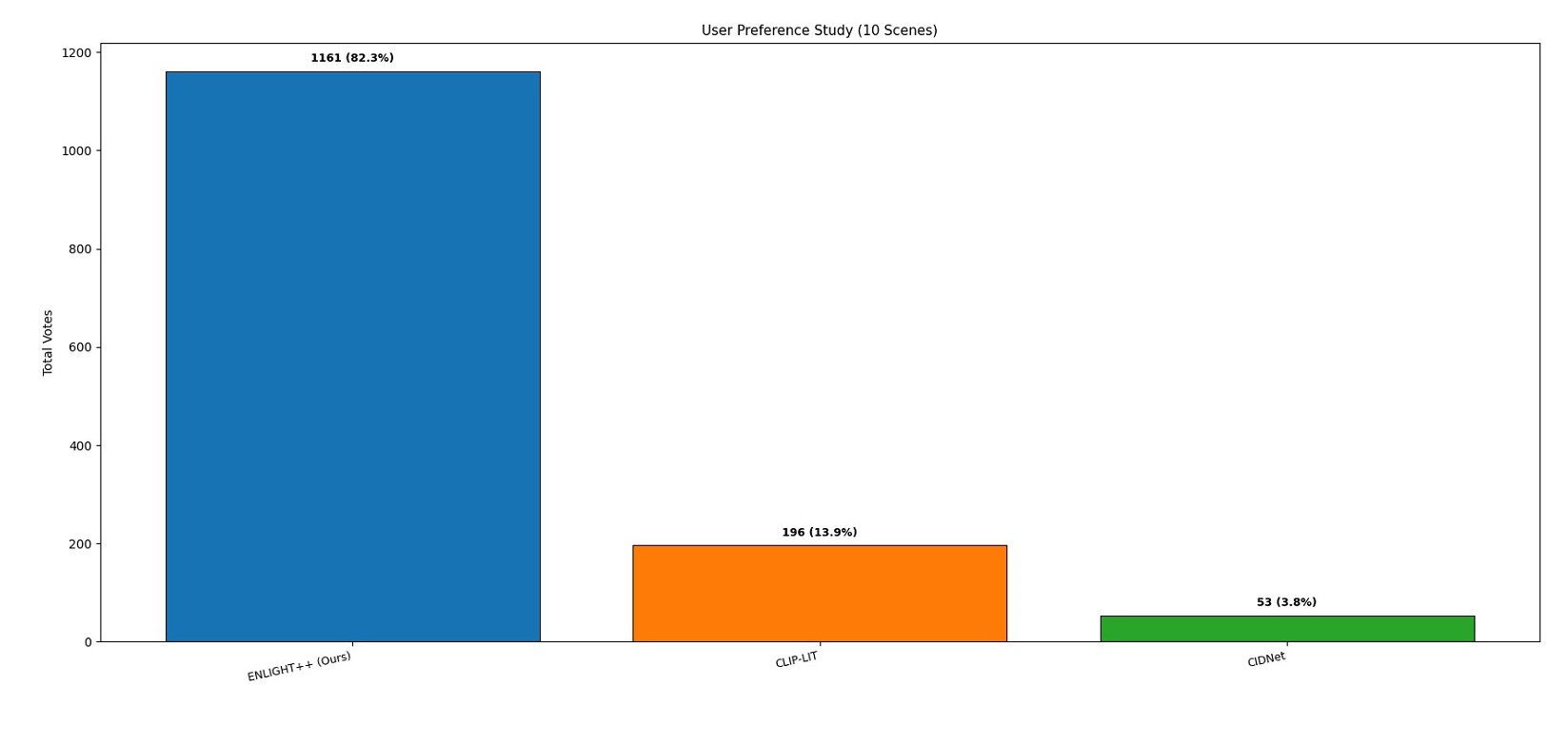}
    \caption{User preference study results across 10 scenes. Bars show total votes with corresponding percentages. Enlight is overwhelmingly preferred over CLIP-LIT\cite{liang2023iterative} and CIDNET\cite{yan2024you}.}
    \label{fig:userstudy_plot}
\end{figure}

As summarized in Figure~\ref{fig:userstudy_plot}, Enlight received
82.3\% of all votes, substantially outperforming CLIP-LIT\cite{liang2023iterative} (13.9\%) and
CIDNET\cite{yan2024you} (3.8\%). In addition, Enlight achieved the highest preference
count in all 10 scenes, indicating a 100\% win rate. Detailed per image statistics can be found at Table \ref{tab:per_image} .
\begin{table}[t]
\centering
\caption{Per-image user preference distribution across the 10 survey scenes.}
\begin{tabular}{c|ccc|c}
\toprule
Image & Enlight & CIDNET\cite{yan2024you} & CLIP-LIT\cite{liang2023iterative} & Winner \\
\midrule
1 & 100 (70.9\%) & 0 (0\%) & 41 (29.1\%) & Ours \\
2 & 130 (92.2\%) & 10 (7.1\%) & 1 (0.7\%) & Ours \\
3 & 108 (76.6\%) & 16 (11.3\%) & 17 (12.1\%) & Ours \\
4 & 120 (85.1\%) & 3 (2.1\%) & 18 (12.8\%) & Ours \\
5 & 109 (77.3\%) & 2 (1.4\%) & 30 (20.1\%) & Ours \\
6 & 130 (92.2\%) & 6 (4.3\%) & 5 (3.5\%) & Ours \\
7 & 94 (66.7\%) & 7 (5.0\%) & 40 (28.4\%) & Ours \\
8 & 119 (84.4\%) & 8 (5.7\%) & 14 (9.9\%) & Ours \\
9 & 139 (98.6\%) & 0 (0\%) & 2 (1.4\%) & Ours \\
10 & 112 (79.4\%) & 1 (0.7\%) & 28 (19.9\%) & Ours \\
\bottomrule
\end{tabular}
\label{tab:per_image}
\end{table}
These findings
demonstrate the strong perceptual alignment of the proposed
proxy-guided enhancement and its consistent competitive performance across diverse
illumination conditions.

\section{Conclusion}

In this work, we introduced \textsc{Enlight}, a training-free and GPU-accelerated optimization framework for low-light image enhancement that does not require paired supervision, pretraining, or task-specific priors. By formulating enhancement as a multi-objective search over entropy, gradient consistency, and noise penalties, the proposed method adapts to diverse lighting conditions through a perceptually guided optimization process.

We further proposed two practical configurations, \emph{Fast} and \emph{Ultrafast}, which provide flexible trade-offs between perceptual quality and computational efficiency. Experiments on five standard unpaired LLIE datasets (DICM, LIME, MEF, NPE, and VV) show that \textsc{Enlight} achieves competitive perceptual quality while maintaining high efficiency. In particular, the \emph{Ultrafast} variant operates within approximately 0.10--0.23 seconds per image, enabling near real-time performance on modern GPUs.

Ablation studies highlight the contribution of warm-start initialization and shadow-aware refinement. Additional evaluations on challenging backlit datasets (Backlit300 and BAID) suggest that the proposed framework generalizes reasonably well to scenes exhibiting strong brightness imbalance. By directly optimizing a perceptually motivated objective, \textsc{Enlight} provides an interpretable alternative to purely learning-based enhancement methods while maintaining a favorable balance between quality and computational cost.

\subsection{Limitations and Future Work}

Despite its promising performance, several limitations should be acknowledged. First, the proposed objective function is primarily motivated by empirical observations and combines entropy, gradient consistency, and noise penalties as perceptual proxies rather than deriving them from a formal model of human visual perception. Second, while the method generally improves no-reference image quality metrics, such metrics do not always perfectly correlate with subjective human preference. Third, some of the design choices, including the shadow threshold used during local refinement and the parameter search ranges, were selected empirically and may not be universally optimal across all imaging conditions. 

Future work will investigate adaptive threshold selection, more principled perceptual objectives and extensions to video enhancement and resource constrained deployment settings.


\FloatBarrier







\bibliographystyle{elsarticle-num}
\bibliography{references}

\begin{thebibliography}{10}
\expandafter\ifx\csname url\endcsname\relax
  \def\url#1{\texttt{#1}}\fi
\expandafter\ifx\csname urlprefix\endcsname\relax\def\urlprefix{URL }\fi
\expandafter\ifx\csname href\endcsname\relax
  \def\href#1#2{#2} \def\path#1{#1}\fi

\bibitem{guo2016lime}
X.~Guo, Y.~Li, H.~Ling, Lime: Low-light image enhancement via illumination map estimation, IEEE Transactions on image processing 26~(2) (2016) 982--993.

\bibitem{guo2020zero}
C.~Guo, C.~Li, J.~Guo, C.~C. Loy, J.~Hou, S.~Kwong, R.~Cong, Zero-reference deep curve estimation for low-light image enhancement, in: Proceedings of the IEEE/CVF conference on computer vision and pattern recognition, 2020, pp. 1780--1789.

\bibitem{zhang2019kindling}
Y.~Zhang, J.~Zhang, X.~Guo, Kindling the darkness: A practical low-light image enhancer, in: Proceedings of the 27th ACM international conference on multimedia, 2019, pp. 1632--1640.

\bibitem{land1971lightness}
E.~H. Land, J.~J. McCann, Lightness and retinex theory, Journal of the Optical society of America 61~(1) (1971) 1--11.

\bibitem{fu2016weighted}
X.~Fu, D.~Zeng, Y.~Huang, X.-P. Zhang, X.~Ding, A weighted variational model for simultaneous reflectance and illumination estimation, in: Proceedings of the IEEE conference on computer vision and pattern recognition, 2016, pp. 2782--2790.

\bibitem{ma2015multi}
K.~Ma, Z.~Wang, Multi-exposure image fusion: A patch-wise approach, in: 2015 IEEE International Conference on Image Processing (ICIP), IEEE, 2015, pp. 1717--1721.

\bibitem{li2018multi}
H.~Li, L.~Zhang, Multi-exposure fusion with cnn features, in: 2018 25th IEEE International Conference on Image Processing (ICIP), IEEE, 2018, pp. 1723--1727.

\bibitem{wei2018deep}
C.~Wei, W.~Wang, W.~Yang, J.~Liu, Deep retinex decomposition for low-light enhancement, arXiv preprint arXiv:1808.04560 (2018).

\bibitem{wu2022uretinex}
W.~Wu, J.~Weng, P.~Zhang, X.~Wang, W.~Yang, J.~Jiang, Uretinex-net: Retinex-based deep unfolding network for low-light image enhancement, in: Proceedings of the IEEE/CVF conference on computer vision and pattern recognition, 2022, pp. 5901--5910.

\bibitem{jiang2021enlightengan}
Y.~Jiang, X.~Gong, D.~Liu, Y.~Cheng, C.~Fang, X.~Shen, J.~Yang, P.~Zhou, Z.~Wang, Enlightengan: Deep light enhancement without paired supervision, IEEE transactions on image processing 30 (2021) 2340--2349.

\bibitem{li2021learning}
C.~Li, C.~Guo, C.~C. Loy, Learning to enhance low-light image via zero-reference deep curve estimation, IEEE transactions on pattern analysis and machine intelligence 44~(8) (2021) 4225--4238.

\bibitem{liu2021retinex}
R.~Liu, L.~Ma, J.~Zhang, X.~Fan, Z.~Luo, Retinex-inspired unrolling with cooperative prior architecture search for low-light image enhancement, in: Proceedings of the IEEE/CVF conference on computer vision and pattern recognition, 2021, pp. 10561--10570.

\bibitem{ma2022toward}
L.~Ma, T.~Ma, R.~Liu, X.~Fan, Z.~Luo, Toward fast, flexible, and robust low-light image enhancement, in: Proceedings of the IEEE/CVF conference on computer vision and pattern recognition, 2022, pp. 5637--5646.

\bibitem{mittal2012no}
A.~Mittal, A.~K. Moorthy, A.~C. Bovik, No-reference image quality assessment in the spatial domain, IEEE Transactions on image processing 21~(12) (2012) 4695--4708.

\bibitem{mittal2013making}
A.~Mittal, R.~Soundararajan, A.~C. Bovik, Making a “completely blind” image quality analyzer.", IEEE Signal Processing Letters 20~(3) (2013) 209--212.
\newblock \href {https://doi.org/10.1109/LSP.2012.2227726} {\path{doi:10.1109/LSP.2012.2227726}}.

\bibitem{kimmel2003variational}
R.~Kimmel, M.~Elad, D.~Shaked, R.~Keshet, I.~Sobel, A variational framework for retinex, International Journal of computer vision 52~(1) (2003) 7--23.

\bibitem{ng2011total}
M.~K. Ng, W.~Wang, A total variation model for retinex, SIAM Journal on Imaging Sciences 4~(1) (2011) 345--365.

\bibitem{zhu2020eemefn}
M.~Zhu, P.~Pan, W.~Chen, Y.~Yang, Eemefn: Low-light image enhancement via edge-enhanced multi-exposure fusion network, in: Proceedings of the AAAI conference on artificial intelligence, Vol.~34, 2020, pp. 13106--13113.

\bibitem{ying2017new}
Z.~Ying, G.~Li, Y.~Ren, R.~Wang, W.~Wang, A new image contrast enhancement algorithm using exposure fusion framework, in: International conference on computer analysis of images and patterns, Springer, 2017, pp. 36--46.

\bibitem{fu2015probabilistic}
X.~Fu, Y.~Liao, D.~Zeng, Y.~Huang, X.-P. Zhang, X.~Ding, A probabilistic method for image enhancement with simultaneous illumination and reflectance estimation, IEEE Transactions on Image Processing 24~(12) (2015) 4965--4977.

\bibitem{lee2013contrast}
C.~Lee, C.~Lee, C.-S. Kim, Contrast enhancement based on layered difference representation of 2d histograms, IEEE transactions on image processing 22~(12) (2013) 5372--5384.

\bibitem{ma2015perceptual}
K.~Ma, K.~Zeng, Z.~Wang, Perceptual quality assessment for multi-exposure image fusion, IEEE Transactions on Image Processing 24~(11) (2015) 3345--3356.

\bibitem{wang2013naturalness}
S.~Wang, J.~Zheng, H.-M. Hu, B.~Li, Naturalness preserved enhancement algorithm for non-uniform illumination images, IEEE transactions on image processing 22~(9) (2013) 3538--3548.

\bibitem{vonikakis2018evaluation}
V.~Vonikakis, R.~Kouskouridas, A.~Gasteratos, On the evaluation of illumination compensation algorithms, Multimedia Tools and Applications 77~(8) (2018) 9211--9231.

\bibitem{liang2023iterative}
Z.~Liang, C.~Li, S.~Zhou, R.~Feng, C.~C. Loy, Iterative prompt learning for unsupervised backlit image enhancement, in: Proceedings of the IEEE/CVF International Conference on Computer Vision, 2023, pp. 8094--8103.

\bibitem{bychkovsky2011learning}
V.~Bychkovsky, S.~Paris, E.~Chan, F.~Durand, Learning photographic global tonal adjustment with a database of input/output image pairs, in: CVPR 2011, IEEE, 2011, pp. 97--104.

\bibitem{wang2022low}
Y.~Wang, R.~Wan, W.~Yang, H.~Li, L.-P. Chau, A.~Kot, Low-light image enhancement with normalizing flow, in: Proceedings of the AAAI conference on artificial intelligence, Vol.~36, 2022, pp. 2604--2612.

\bibitem{xu2022snr}
X.~Xu, R.~Wang, C.-W. Fu, J.~Jia, Snr-aware low-light image enhancement, in: Proceedings of the IEEE/CVF conference on computer vision and pattern recognition, 2022, pp. 17714--17724.

\bibitem{fu2023learning}
Z.~Fu, Y.~Yang, X.~Tu, Y.~Huang, X.~Ding, K.-K. Ma, Learning a simple low-light image enhancer from paired low-light instances, in: Proceedings of the IEEE/CVF conference on computer vision and pattern recognition, 2023, pp. 22252--22261.

\bibitem{yan2024you}
Q.~Yan, Y.~Feng, C.~Zhang, P.~Wang, P.~Wu, W.~Dong, J.~Sun, Y.~Zhang, You only need one color space: An efficient network for low-light image enhancement, arXiv preprint arXiv:2402.05809 (2024).

\bibitem{afifi2021learning}
M.~Afifi, K.~G. Derpanis, B.~Ommer, M.~S. Brown, Learning multi-scale photo exposure correction, in: Proceedings of the IEEE/CVF conference on computer vision and pattern recognition, 2021, pp. 9157--9167.

\bibitem{zhao2021deep}
L.~Zhao, S.-P. Lu, T.~Chen, Z.~Yang, A.~Shamir, Deep symmetric network for underexposed image enhancement with recurrent attentional learning, in: Proceedings of the IEEE/CVF international conference on computer vision, 2021, pp. 12075--12084.

\bibitem{zhang2019zero}
L.~Zhang, L.~Zhang, X.~Liu, Y.~Shen, S.~Zhang, S.~Zhao, Zero-shot restoration of back-lit images using deep internal learning, in: Proceedings of the 27th ACM international conference on multimedia, 2019, pp. 1623--1631.

\end{thebibliography}

\end{document}